\documentclass[10pt,twocolumn,letterpaper]{article}

\usepackage{wacv}              

\definecolor{wacvblue}{rgb}{0.21,0.49,0.74}
\usepackage[pagebackref,breaklinks,colorlinks,allcolors=wacvblue]{hyperref}

\usepackage{enumitem}
\usepackage{multirow}

\def\wacvPaperID{1737} 
\def\confName{WACV}
\def\confYear{2026}

\usepackage{mathtools} 

\title{HyPCA-Net: Advancing Multimodal Fusion in Medical Image Analysis}

\author{
  Joy Dhar\textsuperscript{1,*} \and
  Manish Kumar Pandey\textsuperscript{2,*} \and
  Debashis Das Chakladar\textsuperscript{3,*} \and
  Maryam Haghighat\textsuperscript{4} \and
  Azadeh Alavi\textsuperscript{5} \and
  Sajib Mistry\textsuperscript{6} \and
  Nayyar Zaidi\textsuperscript{7} \and \\
  {\small \textsuperscript{1}Indian Institute of Technology Ropar, India} \quad
  {\small \textsuperscript{2}RoentGen Health, India} \quad
  {\small \textsuperscript{3}Lulea University of Technology, Sweden} \quad \\
  {\small \textsuperscript{4}QUT, Australia} \quad
  {\small \textsuperscript{5}RMIT University, Australia} \quad
  {\small \textsuperscript{6}Curtin University, Australia} \quad
  {\small \textsuperscript{7}Deakin University, Australia}
}

\begin{document}

\twocolumn[{%
  \renewcommand\twocolumn[1][]{#1}%
  \maketitle                       
  \captionsetup{type=figure}       
  \centering
  \includegraphics[width=0.95\linewidth]{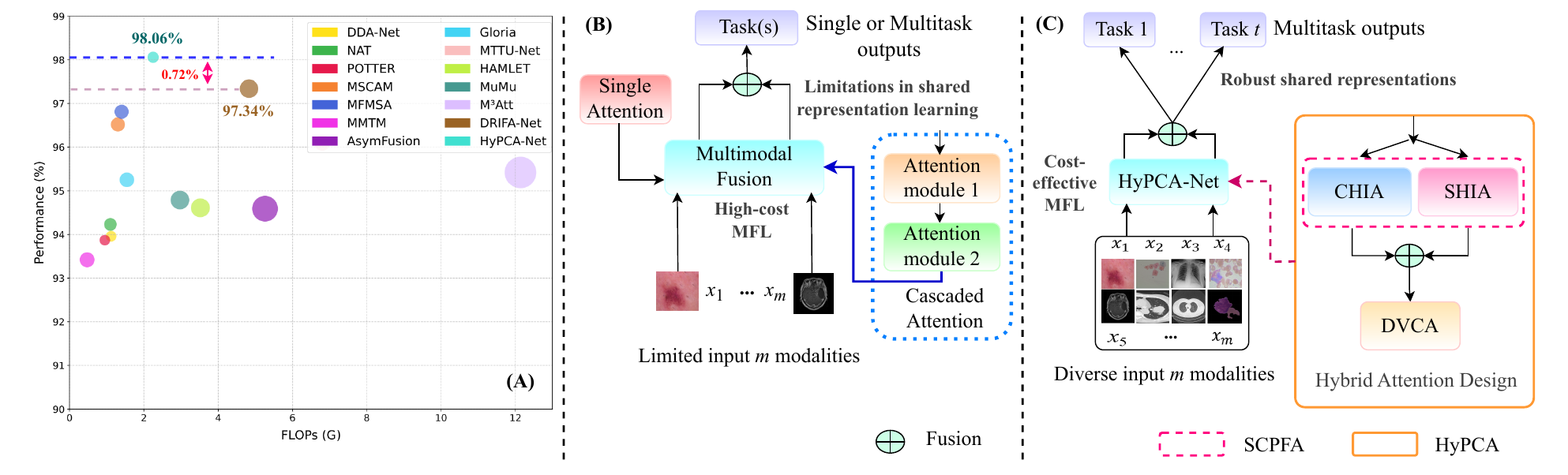}
\caption{\small
(A) Performance comparison between unimodal baselines (e.g., \texttt{DDA-Net \cite{cui2023dual}, NAT \cite{hassani2023neighborhood}, POTTER \cite{zheng2023potter}}, etc.)  and computationally expensive multimodal fusion learning (\texttt{MFL}) methods (e.g., \texttt{DRIFA-Net} \cite{dhar2024multimodal}, 
\texttt{HAMLET} \cite{islam2020hamlet}, and
\texttt{MuMu} \cite{islam2022mumu}, etc.) 
versus our proposed framework -- \texttt{HyPCA-Net}. Bubble size reflects parameter count (larger bubbles = more parameters). (B-C) High-level comparison of three attention mechanisms, namely single attention~\cite{zheng2023potter,hassani2023neighborhood}; cascaded attention~\cite{dhar2024multimodal}; and our proposed hybrid attention (\texttt{HyPCA-Net}). 
(C) An overview of~\texttt{HyPCA-Net}, featuring a \texttt{HyPCA} block that integrates (i) parallel spatial–channel fusion via \texttt{CHIA} and \texttt{SHIA} within the \texttt{SCPFA} module, and (ii) cascaded hybrid-space dual-domain modeling via the \texttt{DVCA} module.}
 \label{fig:fig1}
}
  \vspace{1em}
]


\maketitle

\begingroup
\renewcommand\thefootnote{\fnsymbol{footnote}}
\footnotetext[1]{These authors contributed equally.}
\endgroup

\begin{abstract}

Multimodal fusion frameworks, which integrate diverse medical imaging modalities (e.g., \texttt{MRI}, \texttt{CT}), have shown great potential in applications such as skin cancer detection, dementia diagnosis, and brain tumor prediction. However, existing multimodal fusion methods face significant challenges. First, they often rely on computationally expensive models, limiting their applicability in low-resource environments. Second, they often employ cascaded attention modules, which potentially increase risk of information loss during inter-module transitions and hinder their capacity to effectively capture robust shared representations across modalities. This restricts their generalization in multi-disease analysis tasks.
To address these limitations, we propose a Hybrid Parallel‐Fusion Cascaded Attention Network ($\texttt{HyPCA-Net}$), composed of two core novel blocks: (a) a computationally efficient residual adaptive learning attention block for capturing refined modality-specific representations, and (b) a dual-view cascaded attention block aimed at learning robust shared representations across diverse modalities. Extensive experiments on ten publicly available datasets exhibit that $\texttt{HyPCA-Net}$ significantly outperforms existing leading methods, with improvements of up to 5.2\% in performance and reductions of up to 73.1\% in computational cost. {\color{blue}{Code:~\url{https://github.com/misti1203/HyPCA-Net}.}}


\end{abstract}    



\begin{figure*}[ht!]
    \centering
    \includegraphics[width=\textwidth]{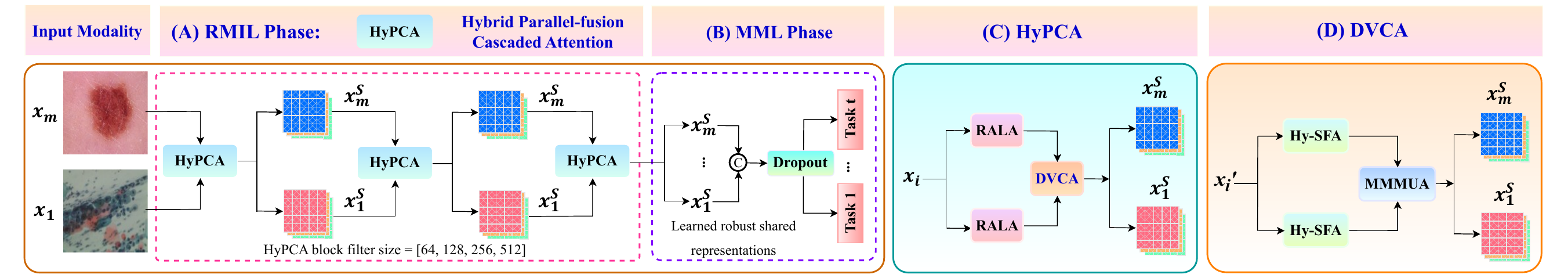}
    \vspace{-0.5cm}
    \caption{\small
Overview of \texttt{HyPCA-Net} framework composing of two phases: (A) \texttt{RMIL}, which learns robust shared representations $\{x_i^{S}\}_{i=1}^{m}$; and (B) \texttt{MML}, which performs multi-disease classification. Within \texttt{RMIL} phase stands our novel (C) \texttt{HyPCA} block comprises of \texttt{RALA} block (refining unimodal features $\{x'_i\}_{i=1}^{m}$) and (D) \texttt{DVCA} block (capturing dual-domain multimodal information)}
    \label{fig:fig2}
    \vspace{-0.4cm}
\end{figure*}

\vspace{-0.5cm}
\section{Introduction} 
\label{sec:introduction}
\vspace{-0.1cm}
Multimodal data is vital in real-world applications, especially in medical image analysis. Imaging modalities like \texttt{MRI}, \texttt{CT}, and \texttt{CXR} offer complementary anatomical and functional information crucial for diagnosing tumors and other pathologies \cite{dhar2024multimodal}. While deep learning has enabled efficient diagnosis, single-modality models even with  feature fusion or attention, often under-perform due to noisy inputs and overfitting issues \cite{dhar2024multimodal, dhar2024uncertainty, cui2023dual, hassani2023neighborhood, zheng2023potter}. 
Multimodal fusion mitigates these issues by learning shared representations across heterogeneous modalities~\cite{chakladar2021multimodal,das2024fusion} (Figure~\ref{fig:fig1} (B)). Attention mechanisms further complement leaning by enabling models to focus on salient regions or biomarkers, hence improving representation learning in various medical imaging tasks \cite{huang2021gloria,islam2020hamlet,islam2022mumu,liu2023multi} (Figure~\ref{fig:fig1} (B)).
Despite recent advances, existing multimodal fusion frameworks \cite{dhar2024multimodal, islam2022mumu, islam2020hamlet, liu2023multi} incur high computational cost (\textbf{challenge 1}) from standard convolutions and multi-stage attention, limiting scalability in healthcare AI. Their sequential (cascaded) attention designs also induce progressive information loss (\textbf{challenge 2}) across inter-module transitions, restricting the retention of discriminative patterns and leading to suboptimal performance \cite{shen2021parallel, lv2024lightweight1}.
These methods do not leverage parallel fusion attention, which can preserve crucial information and enhance representation learning \cite{shen2021parallel, lv2024lightweight1}.

To address these limitations, we propose a \textbf{Hy}brid \textbf{P}arallel-fusion \textbf{C}ascaded \textbf{A}ttention \textbf{Net}work -- \texttt{HyPCA-Net} -- a multimodal fusion framework that balances optimal performance with minimal computational cost, while preserving information through both parallel and cascaded attention (Figure~\ref{fig:fig1} (C)). 
Our proposed framework addresses the aforementioned challenges through two novel blocks: \emph{a Residual Adaptive Learning Attention} (\texttt{RALA}) block and \emph{a Dual-View Cascaded Attention} (\texttt{DVCA}) module, leading to superior performance compared to state-of-the-art methods (Figure~\ref{fig:fig1} (A)). 
Note that \texttt{RALA} and \texttt{DVCA} together constitute our novel \texttt{HyPCA} block, as shown in Figure~\ref{fig:fig2}. Once integrated into a multimodal learning framework comprising two phases—\texttt{RMIL} and \texttt{MML}—we refer to it as \texttt{HyPCA-Net}. Various components of \texttt{HyPCA-Net}, along with the inputs and outputs of each block, are illustrated in Figure~\ref{fig:fig2}.

\texttt{HyPCA-Net} balances optimal performance with minimal computational cost across diverse medical imaging modalities.
Specifically, the \texttt{RALA} block addresses \textbf{both challenges} by learning \textit{enhanced multi-scale spatial–channel representations in parallel} via an efficient Spatial–Channel convolution Adaptive Learning Attention (SCALA) module. This design preserves computational efficiency, enriches representational diversity, and bypasses the sequential bottlenecks of cascaded attention.
On the other hand,~\texttt{DVCA} block \textbf{tackles challenge 1} by leveraging \textit{cascaded hybrid spaces and dual-domain information integration}. By optimizing multimodal features across these spaces and domains, \texttt{DVCA} learns robust shared representations while reducing computational overhead.
By integrating these components, \texttt{HyPCA-Net} is tailored to learn from diverse medical imaging scenarios and exhibits strong generalizability for multi-disease classification. \textbf{Our main contributions can be summarized as follows:}

\vspace{-0.05cm}
\begin{itemize} [leftmargin=*] 
\item[$\bullet$] We propose \texttt{HyPCA-Net}, a multimodal fusion framework that seamlessly integrates parallel fusion attention and cascaded attention modules to learn robust shared representations while achieving optimal performance at minimal computational cost. 
\item[$\bullet$] We propose a novel~\texttt{RALA} block, which enhances unimodal representations by incorporating multi-scale convolutions with a parallel spatial–channel fusion attention module, thereby progressively refining features through the joint capture of diverse spatial and channel dependencies. 
\item[$\bullet$] We introduce a novel \texttt{DVCA} block, a cascaded attention module that integrates dual-domain information within a hybrid representation space to learn robust shared features. 
\item[$\bullet$] We conduct extensive experiments on ten diverse medical imaging datasets to demonstrate that \texttt{HyPCA-Net} consistently outperforms state-of-the-art methods. 
\end{itemize} 
We summarize all abbreviations used throughout the paper in Table~\ref{tab:modules-with-parent} to improve the clarity.

\vspace{-0.3cm}
\begin{table}[htbp]
  \centering\footnotesize
  \caption{List of abbreviations based on modules and components.}
  \vspace{-0.3cm}
  \label{tab:modules-with-parent}
  \scalebox{0.75}{
  \begin{tabular}{@{}l p{0.8\linewidth} l@{}}
    \toprule
    \textbf{Abbreviation} & \textbf{Full Term} & \textbf{Appears in} \\
    \midrule
    HyPCA\text{-}Net & Hybrid Parallel-fusion Cascaded Attention Network & Sec.~\ref{sec:introduction}, \ref{sec:methods}  \\
    HyPCA            & HyPCA block (RALA + DVCA)                         & Sec.~\ref{sec:methods} \\
    RMIL             & Robust Multimodal Information Learning            & Sec.~\ref{sec_MSIL} \\
    RALA             & Residual Adaptive Learning Attention              & Sec.~\ref{rala} \\
    
    SCALA            & Spatial–Channel convolution Adaptive Learning Attention      & Sec.~\ref{rala} \\
    MSHC             & Multi-scale Spatial Heterogeneous Convolution      & Sec.~\ref{rala} \\
    SCPFA            & Spatial–Channel Parallel Fusion Attention          & Sec.~\ref{rala} \\
    SHIA             & Spatial Holistic Information-Learning Attention    & Sec.~\ref{rala} \\
    CHIA             & Channel Holistic Information-Learning Attention    & Sec.~\ref{rala} \\
    DVCA             & Dual-View Cascaded Attention                      & Sec.~\ref{dvca} \\
    Hy\text{-}SFA    & Hybrid Space Fusion Attention                      & Sec.~\ref{dvca} \\
    TFSI             & Token-space Frequency–Spatial Integration          & Sec.~\ref{dvca} \\
    FDCA             & Feature-space Dual-Solver Channel Attention        & Sec.~\ref{dvca} \\
    HCA              & Heterogeneous Channel Attention                    & Sec.~\ref{dvca} \\
    MMMUA            & Multi-scale Multi-frequency Mutual Update Attnention    & Sec.~\ref{dvca} \\
    FCIF             & Frequency-domain Channel Information Fusion        & Sec.~\ref{dvca} \\
    SMIF             & Spatial-domain Multi-scale Information Fusion      & Sec.~\ref{dvca} \\
    MCBI             & Mutual Cross Bidirectional Interactions            & Sec.~\ref{dvca} \\
    HCF              & Hierarchical Channel Fusion                        & Sec.~\ref{dvca} \\
    MML              & Multimodal Multitask Learning                     & Sec.~\ref{sec:methods} \\
    \bottomrule
  \end{tabular}
  }
  \vspace{-0.5cm}
\end{table}


\begin{figure*}[ht!]
    \centering
    \includegraphics[width=0.87\textwidth]{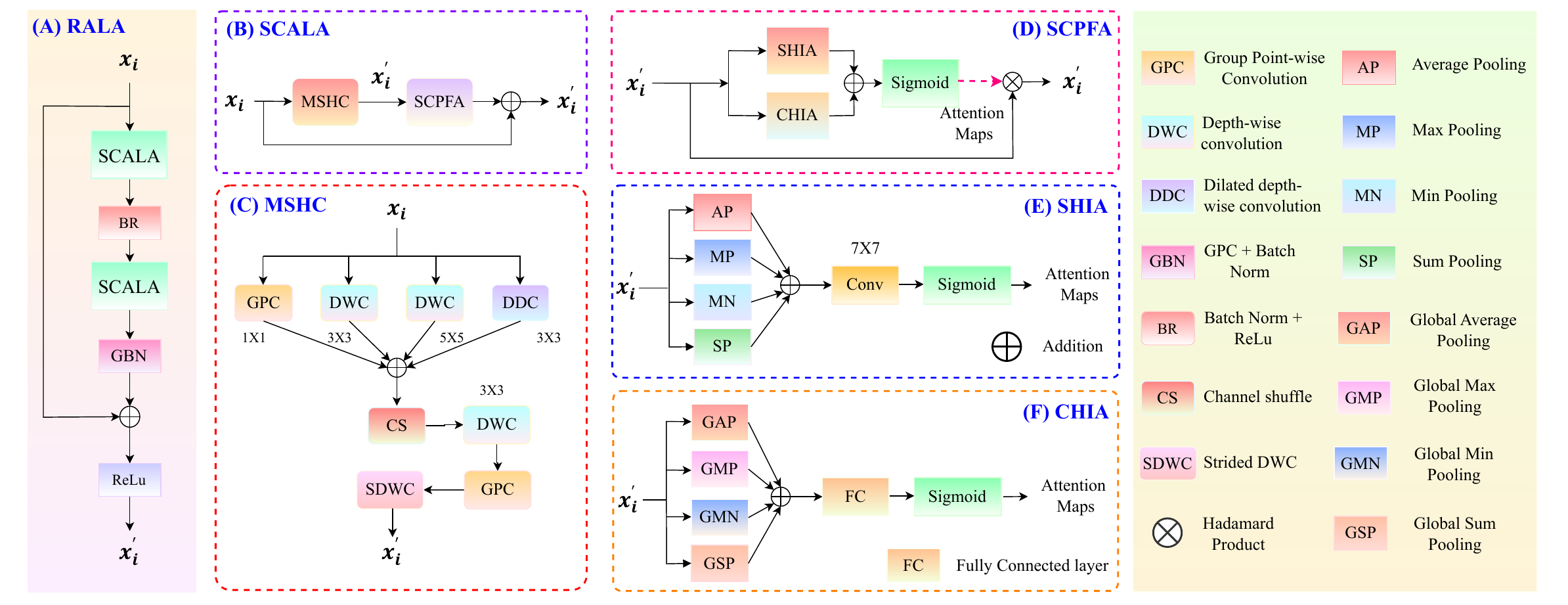}
    \vspace{-0.2cm}
    \caption{\small
(A) Overview of \texttt{RALA} block (first main component of \texttt{HyPCA}), which is 
composed of \texttt{SCALA} blocks (B), which in turn is made out of \texttt{MSHC} block (C) 
and \texttt{SCPFA} block (D). The \texttt{SCPFA} block is composed of
\texttt{SHIA} (E) and \texttt{CHIA} (F) components.}
    \label{fig:fig3}
    \vspace{-0.4cm}
\end{figure*}


\section{Related Works}
\label{sec:related_works}
\vspace{-0.2cm}
Earlier multimodal fusion research has extensively focused on natural vision and medical imaging applications \cite{peng2022balanced, joze2020mmtm, cheng2022fully}. Specifically, existing frameworks e.g., \texttt{MTTU-Net} integrated \texttt{CNNs} and transformers for glioma segmentation and \texttt{IDH} prediction but lacked explicit attention for refined representations \cite{cheng2022fully}. 
Recent attention-driven designs -- \texttt{DDA-Net} \cite{cui2023dual}, \texttt{MADGNet} \cite{nam2024modality}, \texttt{NAT} \cite{hassani2023neighborhood}, \texttt{POTTER} \cite{zheng2023potter}, and \texttt{EMCAD} -- show substantial gains in representation learning \cite{islam2020hamlet, islam2022mumu, huang2021gloria, He2023, liu2023multi}. Incorporating attention mechanisms into multimodal fusion frameworks -- through structural designs such as parallel and cascaded attention modules -- has significantly enhanced performance \cite{islam2020hamlet, islam2022mumu, huang2021gloria, He2023, liu2023multi}. 
For example, \texttt{Gloria} integrated global–local attention to align radiology images with textual reports for efficient label learning \cite{huang2021gloria}, while \texttt{M\textsuperscript{3}Att} employed mutual attention and iterative interaction to fuse visual and language features for referring image segmentation \cite{liu2023multi}. \texttt{HAMLET} and \texttt{MuMu} utilized hierarchical and guided multimodal attentions for human activity recognition \cite{islam2020hamlet, islam2022mumu}. Methods such as \texttt{CAF} and \texttt{DRIFA-Net} used co-attention fusion and cascaded dual-attention strategies for skin cancer diagnosis and multi-disease classification \cite{He2023, dhar2024multimodal}, showing attention’s versatility.
Despite recent progress, prior frameworks remain computationally intensive and prone to information loss across modules, limiting robust representation learning. To overcome this, we propose a hybrid attention-based fusion architecture 
to balance performance and efficiency, key to scalable healthcare AI. \emph{See \textcolor{blue}{supplementary} for more details.}

\section{Methods} \label{sec:methods}
\textbf{Problem Formulation.} 
Given inputs $X = \{x_i\}_{i=1}^{m}$ drawn from $m$ heterogeneous modalities and labels $Y = \{y_j\}_{j=1}^{t}$ for $t$ tasks, our network $\mathcal{F}(\cdot)$ aims to learn robust shared representations $X^{S} = \{x_i^{s}\}_{i=1}^{m}$. This facilitates the mapping $\mathcal{F}(X) \to Y$, thereby jointly optimizing for high performance with minimal computational cost.


\noindent\textbf{Method Overview.} We present a holistic overview of \texttt{HyPCA-Net} framework (ref. Figure~\ref{fig:fig2}). 
\texttt{HyPCA-Net} comprises of two salient phases:
\textbf{1) Robust Multimodal Information Learning (RMIL)} -- 
Learn robust shared representations. \textbf{2) Multimodal Multitask Learning (MML)} -- inspired from~\texttt{DRIFA-Net}~\cite{dhar2024multimodal} -- it facilitates multiple downstream tasks (e.g., multi-disease classification) and comes after~\texttt{RMIL} phase.
That is, the resulting shared representations \( X^s \) from the \texttt{RMIL} phase is used to classify multiple diseases across \( i \in [1:m] \) modalities. It maps input \( X \) to output predictions \( \mathcal{Y} \) using a loss function \( \mathcal{L}_{\texttt{MML}} \), where \( \lambda_t^i \) controls the weight of each task-modality-specific loss \( \mathcal{L}_t^i \). 
The best model parameters \( \beta^* \) are found by minimizing \( \mathcal{L}_{\texttt{MML}} \), i.e., 

\begin{footnotesize}
\vspace{-0.3cm}
\begin{equation} \label{eq:eq21}
\mathcal{L}_{\texttt{MML}} = \sum_{t=1}^{T} \sum_{i=1}^{m} \lambda_t^i \cdot \mathcal{L}_t^i \big( \mathcal{F}(X^s; \beta), \mathcal{Y} \big); \ \beta^* = \arg \min_\beta  \left(\mathcal{L}_{\texttt{MML}})\right.,
\end{equation}
\end{footnotesize}
\noindent where $\beta$ signifies the \texttt{HyPCA-Net} parameters. 
In the following, we will discuss~\texttt{RMIL} phase in detail.

\noindent\subsection{Robust Multimodal Information Learning} \label{sec_MSIL} 
The Robust Multimodal Information Learning (\texttt{RMIL}) phase 
is comprised of \(m\) parallel modality-specific branches each containing \(b\) cascaded \texttt{HyPCA} blocks
(see Figure \ref{fig:fig2}). 
Each block integrates two attention sub-blocks into a hybrid design by incorporating parallel fusion attention (via \textbf{Residual Adaptive Learning Attention (\texttt{RALA})} block which captures multi-scale spatial and channel dependencies to learn refined unimodal representations $X' = \{\,x'_i\}_{i=1}^m$) with cascaded attention (via \textbf{Dual-View Cascaded Attention (\texttt{DVCA})} block which integrates spatial and frequency domain information to capture robust shared representations $X^s = \{\,x^s_i\}_{i=1}^m$). 
Let us delve into the details of these blocks in the following.

\subsubsection{\textbf{Residual Adaptive Learning Attention}} \label{rala}

Residual Adaptive Learning Attention (\texttt{RALA}) sub-block (ref. Figure~\ref{fig:fig3}(A)) refines representations by jointly capturing multi-scale spatial patterns and channel dependencies with minimal computational cost. 
It is motivated to address \texttt{EMCAD}~\cite{rahman2024emcad} shortcomings,
i.e, 
(1) single-stage multi-scale processing, 
(2) homogeneous branch design, 
and (3) lack of spatial–channel fusion.
\texttt{RALA} addresses these limitations by leveraging  a component named Spatial–Channel convolution Adaptive Learning Attention (\texttt{SCALA}) (Figure~\ref{fig:fig3}(B)) -- which comprises of two core components: A) Multi-scale Spatial Heterogeneous Convolution (\texttt{MSHC}) and B) Spatial-Channel Parallel Fusion Attention (\texttt{SCPFA}).
Specifically to achieve cascaded information refinement: \texttt{RALA} replaces the \texttt{MSDC} block of \texttt{EMCAD} with a pair of \texttt{SCALA} modules (ref. Figure~\ref{fig:fig3}(A)) 
enabling progressive multi-scale representations refinement. 
Secondly, to achieve heterogeneous branch design, each \texttt{MSHC} module employs heterogeneous convolutions at varied scales across branches 
to promote representational diversity. 
Finally, to achieve spatial-channel fusion, an \texttt{SCPFA} block jointly learns spatial and channel dependencies 
resulting enriched representations for downstream tasks. 
\noindent Formally, given a modality-specific input \(x_i\) -- to obtain refine representations $\{x^{\prime}_i\}_{i=1}^m$ -- $\texttt{RALA}(\cdot)$ block can be written as:
\newlength{\lhsw}
\settowidth{\lhsw}{$\texttt{SCALA}(x_i)$} 

\begin{footnotesize}
\vspace{-0.3cm}
\begin{equation}\label{eq:eq1}
\begin{alignedat}{2}
&\mathmakebox[\lhsw][l]{\texttt{RALA}(x_i)} &{}={}&
  \mathcal{R}\Bigl(
    x_i
    + \underbrace{\texttt{BNPC}\bigl(\texttt{SCALA}\bigl(\mathcal{R}(\texttt{BN}(\texttt{SCALA}(x_i)))\bigr)\bigr)}_{\text{Progressive Cascaded Refinement}}
  \Bigr),\\
&\mathmakebox[\lhsw][l]{\texttt{SCALA}(x_i)} &{}={}&
  x_i + \texttt{SCPFA}\bigl(\texttt{MSHC}(x_i)\bigr).
\end{alignedat}
\end{equation}
\end{footnotesize}
where skip connections stabilize gradient flow and promote feature reuse;  $\texttt{BNPC}(\cdot)=\texttt{BN}(\texttt{PC}(\cdot))$ applies batch normalization with point-wise convolution respectively; and $\mathcal{R}(\cdot)$ denotes the \texttt{ReLU} activation.  
In the following, let us discuss~\texttt{MSHC} and~\texttt{SCPFA} blocks. 


\begin{figure}[t]
    \centering
    \includegraphics[width=0.48\textwidth, ]{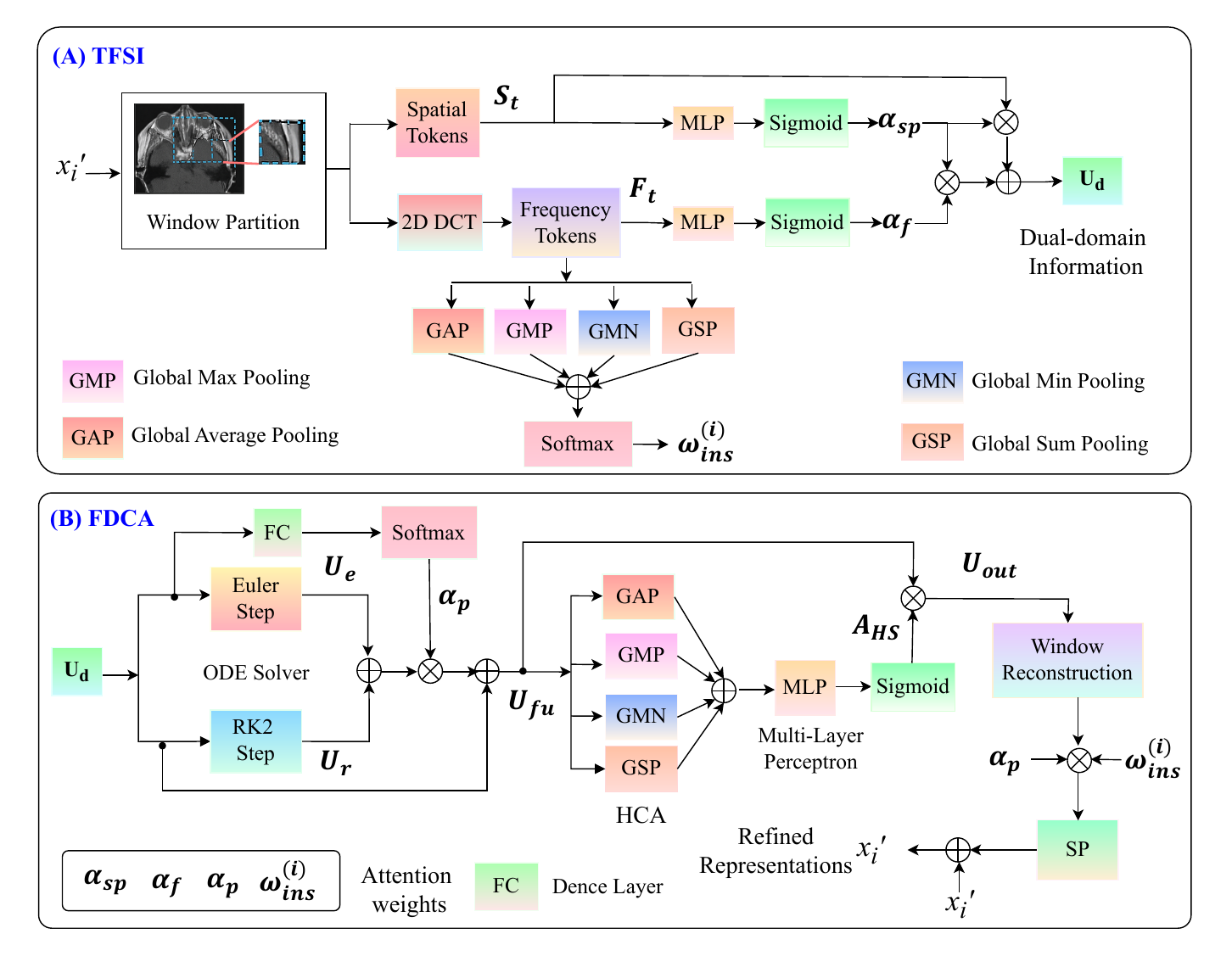}
    \vspace{-0.6cm}
    \caption{\small 
    Overview of \texttt{Hy-SFA} (key component of \texttt{DVCA} block) comprising 
     (A) \texttt{TFSI} and 
    (B) \texttt{FDCA} blocks. \texttt{FDCA} includes Heterogeneous Channel Attention (\texttt{HCA}).}
    \label{fig:fig4}
    \vspace{-0.4cm}
\end{figure}

\noindent\textbf{(A) Multi-scale Spatial Heterogeneous Convolution.} Multi-scale Spatial Heterogeneous Convolution (\texttt{MSHC}) block (ref. Figure~\ref{fig:fig3}(C)) 
employs heterogeneous convolutional branches obtained with group point-wise convolution (\texttt{GPC}), dilated depth-wise convolution (\texttt{DDC}) and depth-wise convolution (\texttt{DWC}) -- with varying scales $k$ (\(1\!\times\!1\), \(3\!\times\!3\), \(5\!\times\!5\)) \footnote{This design fosters branch-wise heterogeneity, enabling the block to capture fine-grained multi-scale spatial patterns by fusing the outputs from these branches to capture enriched representations \( x^{\prime}_i = \texttt{MSHC}(\cdot) \), thus enhancing representational diversity.}. 
To further facilitate inter-channel communication, we apply a channel shuffle operation $\texttt{CS}(\cdot)$. Finally, a sequence of \texttt{GPC}, \texttt{DWC}, and a strided \texttt{DWC} (denoted as \texttt{SDWC}) layers restores the original channel dimensionality, explicitly capturing inter-channel dependencies, and further refining the representation. We summarize operations of~$\texttt{MSHC}(\cdot)$ block as:

\begin{footnotesize}
\begin{equation}\label{eq:eq2}
\texttt{MSHC}(x_i)
= \texttt{SDWC}\Bigl(
    \texttt{GPC}\bigl(
        \texttt{DWC}\bigl(
            \texttt{CS}\bigl(
                \theta\bigl(
                    \underbrace{\forall_{k\in\{1,3,5\}}\texttt{HC}_{k}(x_i)}_{\text{Branch-wise Heterogeneity}}
                \bigr)
            \bigr)
        \bigr)
    \bigr)
\Bigr).
\end{equation}
\end{footnotesize}
where $\texttt{HC} \in \{ \texttt{GPC}, \texttt{DWC}, \texttt{DDC}\}$ and $\theta$ represents fusion.

\noindent\textbf{(B) Spatial–Channel Parallel Fusion Attention.} Spatial–Channel Parallel Fusion Attention (\texttt{SCPFA}) block (ref. Figure~\ref{fig:fig3}(D)) 
enriches modality-specific representations by jointly learning spatial and channel cues. It comprises of two key components: (1) \textit{Channel Holistic Information-Learning Attention} (\texttt{CHIA}) and (2) \textit{Spatial Holistic Information-Learning Attention} (\texttt{SHIA}).
To design \texttt{CHIA} (Figure~\ref{fig:fig3}(F))
, we apply diverse global pooling operations -- global average pooling (\texttt{GAP}), global max pooling (\texttt{GMP}), global min pooling (\texttt{GMN}), and global sum pooling (\texttt{GSP}) -- to capture long-range channel contexts. 
To design \texttt{SHIA} (Figure~\ref{fig:fig3}(E))
, we employ multi-perspective local pooling operations -- average pooling (\texttt{AP}), max pooling (\texttt{MP}), min pooling (\texttt{MN}), and sum pooling (\texttt{SP}) -- to learn fine-grained spatial details. 
Within each branch, the resulting distinct contexts are fused to capture diverse dependencies, i.e., they apply sigmoid $\sigma(\cdot)$ function to obtain the channel attention map $A_C = \texttt{CHIA}(\cdot)$ and the spatial attention map $A_S = \texttt{SHIA}(\cdot)$.  
We integrate these maps into a joint attention map $A_{SC} = \texttt{SCPFA}(\cdot)$, which recalibrate the input features $x'_i$, yielding the refined representations. We write~$\texttt{SCPFA}(\cdot)$ block operations as:

\begin{footnotesize}
\begin{equation}
\begin{aligned}
x^{\prime}_i &= x^{\prime}_i \;\odot\; \texttt{SCPFA}(x^{\prime}_i), \\
\texttt{SCPFA}(x^{\prime}_i) &= \sigma\bigl(\texttt{SHIA}(x^{\prime}_i) + \texttt{CHIA}(x^{\prime}_i)\bigr).
\end{aligned}
\vspace{-0cm}
\end{equation}
\end{footnotesize}
where:
\begin{footnotesize}
\begin{equation}
\texttt{CHIA}(x^{\prime}_i)
= \sigma\Bigl(
    \texttt{FC}\bigl(
        \texttt{GAP}(x^{\prime}_i)
      + \texttt{GMP}(x^{\prime}_i)
      + \texttt{GMN}(x^{\prime}_i)
      + \texttt{GSP}(x^{\prime}_i)
    \bigr)
\Bigr).
\vspace{-0.0cm}
\end{equation}
\end{footnotesize}
\vspace{-0.2cm}
\begin{footnotesize}
\begin{equation}
\texttt{SHIA}(x^{\prime}_i)
= \sigma\Bigl(
    \texttt{Conv}\bigl(
        \texttt{AP}(x^{\prime}_i)
      + \texttt{MP}(x^{\prime}_i)
      + \texttt{MN}(x^{\prime}_i)
      + \texttt{SP}(x^{\prime}_i)
    \bigr)
\Bigr).
\vspace{-0.1cm}
\end{equation}
\end{footnotesize}
Here $\odot$ denotes Hadamard product, \texttt{FC} and \texttt{Conv} are dense and convolution layers, respectively. 


\begin{figure*}[t]
    \centering
    \includegraphics[width=0.813\textwidth]{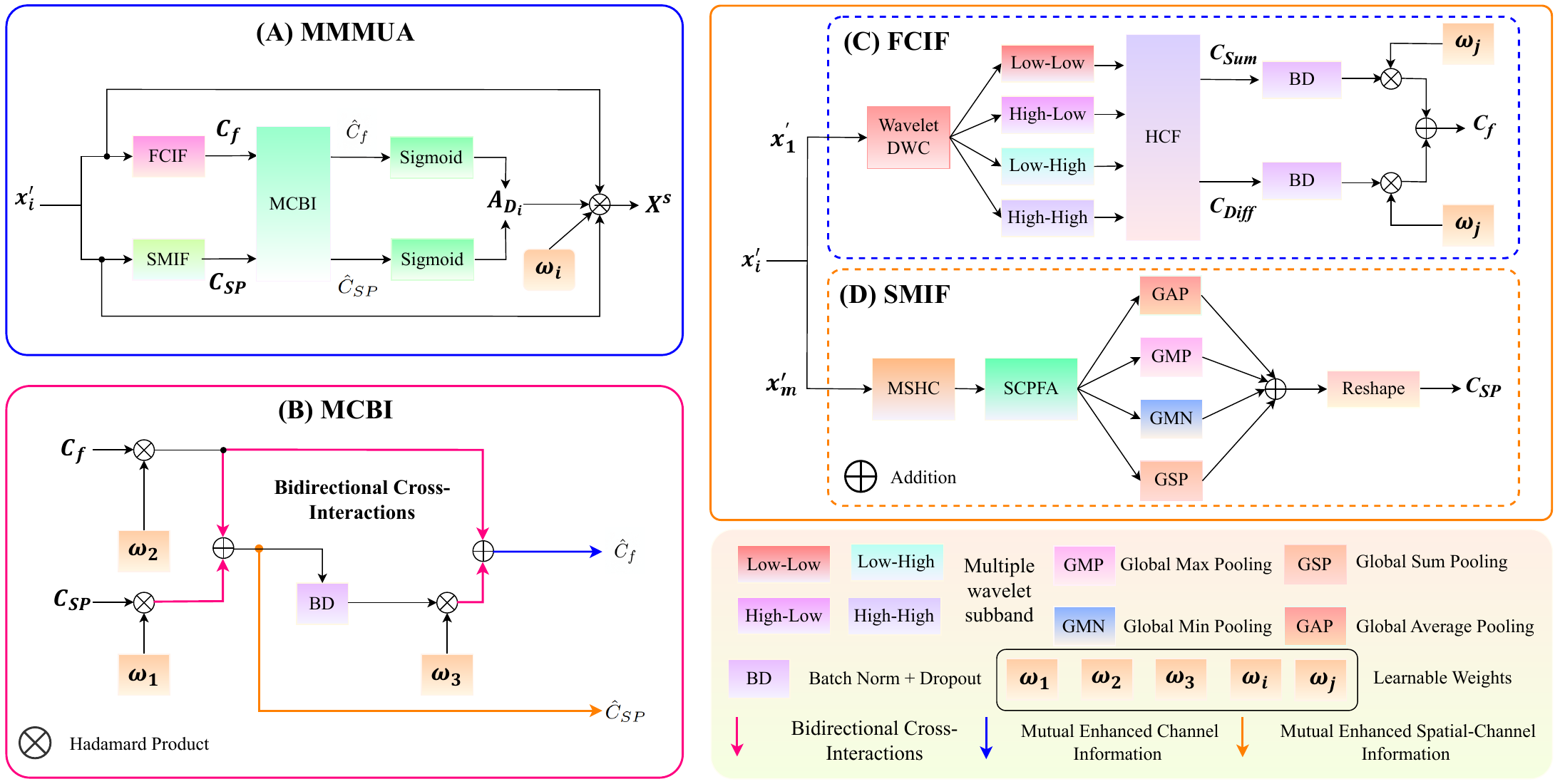}
    \vspace{-0.25cm}
    \caption{\small
\textbf{(A)} Overview of the \texttt{MMMUA} block (key component of \texttt{DVCA} block). 
It consists of \texttt{MCBI} (B), \texttt{FCIF} (C),
and \texttt{SMIF} (D) blocks.  
The \texttt{MCBI} block integrates the outputs of \texttt{FCIF} and \texttt{SMIF} blocks. \texttt{FCIF} employs Hierarchical Channel Fusion (\texttt{HCF}) mechanism.
}
    \label{fig:fig5}
    \vspace{-0.4cm}
\end{figure*}


\subsubsection{\textbf{Dual-View Cascaded Attention}} \label{dvca} 
Dual-View Cascaded Attention (\texttt{DVCA}) block (Figure \ref{fig:fig2}(D)) 
aims to learn robust shared representations \(X^{s}\) from multimodal input \(X^{\prime}\) via two cascaded modules, i.e., Hybrid Space Fusion Attention (\texttt{Hy-SFA}) -- for capturing modality-specific diverse channel dependencies in hybrid (token and feature) spaces and Multi-scale Multi-frequency Mutual Update Attention (\texttt{MMFU}) -- for enriching multimodal contextual representations by capturing dual domain (spatial and frequency) information. Let us discuss these two blocks in the following. 
\noindent\textbf{(A) Hybrid Space Fusion Attention.} Token-space attention (e.g., window-based local attention)~\cite{Liu_2021_ICCV} captures fine-grained spatial interactions, while feature-space method (e.g., frequency-domain attention via \texttt{2D DCT})~\cite{Qin_2021_ICCV} strengthens texture discrimination. Yet these paradigms have not been co-attended within a single module, leaving a gap in hybrid-space interaction for multimodal fusion pipelines. To bridge this gap, we propose the Hybrid Space Fusion Attention (\texttt{Hy-SFA}) module (Figure~\ref{fig:fig4}), which enriches each modality input $x^{\prime}_i$ by jointly capturing diverse channel dependencies across token and feature spaces through two cascaded submodules—\texttt{TFSI} and \texttt{FDCA}—that enable cross-domain context exchange.

\begin{itemize}

\item \emph{Token-space Frequency–Spatial Integration} (\texttt{TFSI}) learns complementary dual-domain information by jointly processing spatial and frequency tokens within each local window (Figure \ref{fig:fig4}(A)). Specifically, given an input $x'_i$ and a window size $w$, we first cyclically shift $x'_i$ by $\lfloor w/4\rfloor$ pixels along spatial axes and partition it into non-overlapping $w\times w$ windows. In each window, \texttt{TFSI} captures spatial tokens $\mathbf{S}_t\in\mathbb{R}^{w^2\times C}$ directly from the windowed features and frequency tokens $\mathbf{F}_t\in\mathbb{R}^{w^2\times C}$ via a normalized \texttt{2D} discrete cosine transform (\texttt{DCT}). Each token set is processed independently via an \texttt{MLP} ($f_{\theta}$) followed by a sigmoid $\sigma$, yielding attention weights:

\vspace{-0.2cm}
\begin{footnotesize}
\begin{equation}
\alpha_{sp} = \sigma\bigl(f_{\theta}(\mathbf{S}_t)\bigr) \quad \text{and} \quad
\alpha_f = \sigma\bigl(f_{\theta}(\mathbf{F}_t)\bigr).
\end{equation}
\end{footnotesize}
We fuse the weighted token streams by emphasizing informative features from each domain to obtain the dual‐domain output $\mathbf{U}_d$:

\vspace{-0.2cm}
\begin{footnotesize}
\begin{equation}
\mathbf{U}_d = \bigl(\alpha_{sp} \odot \alpha_f \bigr) + \bigl(1-\alpha_{sp} \bigr) \odot \mathbf{S}_t.
\end{equation}
\end{footnotesize}


\item \emph{Feature-space Dual-Solver Channel Attention} (\texttt{FDCA}) (Figure \ref{fig:fig4} (B)) refines dual-domain tokens $\mathbf{U}_d \in \mathbb{R}^{HW\times C}$ in feature space by treating them as the initial state ($\mathbf{H}_{0} $) of a continuous‐depth flow, following the neural-ODE formulation \cite{Chen2018NeuralODE}.
A \texttt{GAP} followed by a sigmoid-activated \texttt{Dense} layer applied to $\mathbf{H}_0$, yielding the adaptive step size $\tau\in(0,1]$ for further processing.
To capture information at complementary scales, \texttt{FDCA} applies two discrete Ordinary Differential Equation (\texttt{ODE}) solvers in parallel -- a first-order Euler step for coarse-grained contexts ($\mathbf{U}_{e}$) and a second-order Runge–Kutta (\texttt{RK2}) step for fine-grained details ($\mathbf{U}_{r}$) as:

\vspace{-0.3cm}
\begin{footnotesize}
\begin{equation}
\begin{aligned}
\mathbf{U}_{e} &= \mathbf{H}_{0} + \tau\,f_{\theta}(\mathbf{H}_{0}),\\
k_1 &= f_{\theta}(\mathbf{H}_{0}),\quad
k_2 = f_{\theta}\!\Bigl(\mathbf{H}_{0} + \tfrac{\tau}{2}k_1\Bigr),\quad
\mathbf{U}_{r} = \mathbf{H}_{0} + \tau\,k_2.
\end{aligned}
\vspace{-0.1cm}
\end{equation}
\end{footnotesize}
where $k_1, k_2$ denote intermediate slopes in the \texttt{RK2} update.
The resulting proposals are adaptively fused with attention weights $\alpha_p$ -- generated by a \texttt{Dense}$\rightarrow$\texttt{softmax} layer -- that emphasize the more informative update, while a skip connection preserves the original features $\mathbf{H}_0$, thereby producing refined representations $\mathbf{U}_{\texttt{fu}}$ that jointly capture coarse- and fine-grained details as:

\vspace{-0.2cm}
\begin{footnotesize}
\begin{equation}
\mathbf{U}_{\texttt{fu}} = \alpha_p \bigl(\mathbf{U}_{e} + \mathbf{U}_{r}\bigr) + (1 - \alpha_p)\,\mathbf{H}_0.
\end{equation}
\end{footnotesize}
We achieve Heterogeneous Channel Attention ($\texttt{HCA}$) (Figure \ref{fig:fig4} (B)) by applying diverse global pooling operations -- $\texttt{GAP}$, $\texttt{GMP}$, $\texttt{GMN}$, and $\texttt{GSP}$ -- to the resulting details ($\mathbf{U}_{\texttt{fu}}$) to capture diverse channel statistics. We then aggregate the resulting contexts and feed them via a $\texttt{MLP}$ followed by sigmoid, generating a hybrid space attention map ($A_{\text{HS}}$), which is used to reweight channels according to its learned importance to learn enhanced features:

\vspace{-0.3cm}
\begin{footnotesize}
\begin{equation}
A_{\mathrm{HS}} = \sigma\left(f_\theta\left(\texttt{GAP}(\mathbf{U}_{\texttt{fu}})+\texttt{GMP}(\mathbf{U}_{\texttt{fu}})+\texttt{GMN}(\mathbf{U}_{\texttt{fu}}) + \texttt{GSP}(\mathbf{U}_{\texttt{fu}})\right)\right)
\end{equation}

\end{footnotesize}

\vspace{-0.3cm}
\begin{footnotesize}
\begin{equation}
\mathbf{U}_{\mathrm{out}} = \mathbf{U}_{\mathrm{fu}} \;\odot\; A_{\mathrm{HS}}.
\end{equation}
\end{footnotesize}
The output $\mathbf{U}_{\mathrm{out}}$ undergoes cyclic shift reversal and window reconstruction to original spatial positions, yielding $\tilde{x}_{i,ws}$ for each window scale $ws$ (Figure \ref{fig:fig3} (B)). In parallel, the same global pooling operations are applied to frequency-domain tokens $\mathbf{F}_t$ across all windows, followed by fusing them to learn diverse channel descriptors ($\mathbf{C_e}$) (Figure \ref{fig:fig3} (A)):

\vspace{-0.2cm}
\begin{footnotesize}
\begin{equation}
\mathbf{C_e} = \texttt{GAP}(\mathbf{F}_t) + \texttt{GMP}(\mathbf{F}_t) + \texttt{GMN}(\mathbf{F}_t) + \texttt{GSP}(\mathbf{F}_t).
\end{equation}
\end{footnotesize}
We compute the instance‐specific attention weights i.e., $\omega^{(i)}_{\mathrm{ins}}$ and $\omega_{\mathrm{g}}$ on $\mathbf{C_e}$ and $\alpha_p$, respectively, and then assign them to \(\tilde{x}_{i,ws}\) and leverage $\texttt{SP}$ layer with original input $x'_{i}$, resulting the refined representations:

\vspace{-0.2cm}
\begin{footnotesize}
\begin{equation}
\omega^{(i)}_{\mathrm{ins}} = \mathrm{softmax}(\mathbf{C_e}), \quad
\omega_{\mathrm{g}} = \alpha_p.
\vspace{-0.1cm}
\end{equation}
\end{footnotesize}

\vspace{-0.1cm}
\begin{footnotesize}
\begin{equation}
x'_{i} = x'_{i} + \texttt{SP}\Big(\sum_{ws=1}^{2} \bigl(w^{(i)}_{\mathrm{ins},ws}\,w_{\mathrm{g},ws}\bigr)\,\tilde{x}_{i,ws}\Big)
\end{equation}.
\end{footnotesize}
\end{itemize}


\begin{table*}[htbp]
\centering
\footnotesize
\caption{\small Performance comparison of $\texttt{HyPCA-Net}$ with SOTA methods ($\texttt{M1-M8}$) on $8$ datasets ($\texttt{D1-D8}$) for classification. Bold/underlined values indicate the best/second-best results, respectively. We report no.\ of parameters (in millions) and $\texttt{GFLOPS}$for each method.
}
\vspace{-0.3cm}
\label{tab:tab1}
\scalebox{0.797}{
\setlength{\tabcolsep}{1.68pt}%
\renewcommand{\arraystretch}{1.1}%
\setlength{\doublerulesep}{0.0ex}
\newcommand{\approachsep}{\noalign{\vskip 0.1ex}\hline\hline\noalign{\vskip 0.1ex}}
\begin{tabular}
{l|l|ccc|ccc|ccc|ccc|ccc|ccc|ccc|ccc|cc}
\hline
\multirow{1}{*}{Datasets $\rightarrow$} & \multirow{2}{*}{Backbone} & \multicolumn{3}{c|}{D1: Nickparvar} & \multicolumn{3}{c|}{D2: IQ-OTHNCCD} & \multicolumn{3}{c|}{D3: Tuberculosis} & \multicolumn{3}{c|}{D4: CNMC-2019} & \multicolumn{3}{c|}{D5: HAM10000} & \multicolumn{3}{c|}{D6: SIPaKMeD} & \multicolumn{3}{c|}{D7: CRC}
& \multicolumn{3}{c|}{D8: CBIS-DDSM}
& \multicolumn{2}{c}{Overall}
\\ \cline{1-1} \cline{3-28}
 Models $\downarrow$ & & \multicolumn{1}{c}{ACC} & F1 & AUC & \multicolumn{1}{c}{ACC} & F1 & AUC & \multicolumn{1}{c}{ACC} & F1 & AUC & \multicolumn{1}{c}{ACC} & F1 & AUC & \multicolumn{1}{c}{ACC} & F1 & AUC  & \multicolumn{1}{c}{ACC} & F1 & \multicolumn{1}{c|}{AUC}
 & \multicolumn{1}{c}{ACC} & F1 & \multicolumn{1}{c|}{AUC}
 & \multicolumn{1}{c}{ACC} & F1 & \multicolumn{1}{c|}{AUC} 
& \multicolumn{1}{c}{$\#$P}
& \multicolumn{1}{c}{$\#$F}\\
 \hline
 \multirow{1}{*}{DDA-Net} & ResNet18 & 96.8 & 96.6 & 97.2 & 98.1 & 97.5 & 98.2 & 
 95.8 & 95.1 & 95.4 & 
 92.2 & 91.8 & 92.1 & 
 92.2 & 92.0 & 92.6 & 
 92.5 & 92.8 & 91.9 &
 92.7 & 92.7 & 93.1 &
 92.5 & 91.8 & 91.5 &
$\textbf{12.1}$ & $\textbf{1.12}$ \\
\multirow{1}{*}{MSCAM}  & PVT2-B2 & 97.6 & 97.6 & 97.9 
& 99.5 & 99.5 & 99.5 & 
97.8 & 96.9 & 97.6 
& 95.3 & 95.1 & 95.3 & 
97.6 & 97.2 & 97.6 & 
94.2 & 94.0 & 94.9 & 
96.4 & 96.2 & 96.6 
& 94.1 & 93.9 & 94.2 
& 26.9 & $\underline{1.3}$ \\
\multirow{1}{*}{MFMSA}  & ResNet50 & 97.9 & 97.7 & 98.0 
& 99.5 & 99.3 & 99.5 & 
98.1 & 97.3 & 98.3
& 95.6 & 95.2 & 95.9 & 
97.9 & 97.4 & 97.9 & 
94.8 & 94.4 & 95.3 & 
96.7 & 96.5 & 96.6 
& 94.6 & 94.5 & 94.6 
& 26.9 & 1.4 \\
\approachsep
\multirow{1}{*}{Gloria}  & ResNet50 & 98.1 & 97.6 & 97.9 
& 98.5 & 98.4 & 98.5
& 96.6 & 96.0 & 96.9 
& 93.3 & 93.3 & 93.4 
& 93.8 & 93.8 & 94.5 
& 94.2 & 94.2 & 94.2
& 95.9 & 95.6 & 95.7 
& 92.5 & 91.2 & 91.9 
& 30.8 & 1.54  \\
\multirow{1}{*}{MTTU-Net}  & ResNet50 & 97.9 & 97.9 & 98.0 & 99.5 & 99.2 & 99.5 
& 97.3 & 96.6 & 97.6 
& 94.3 & 93.9 & 94.1 
& 97.4 & 96.5 & 97.2 
& 91.9 & 92.3 & 92.6 
& 96.9 & 96.8 & 97.0 
& 94.1 & 93.3 & 94.5 
& 38.1 & 6.8   \\
\multirow{1}{*}{MuMu} & ResNet50 & 96.8 & 96.8 & 97.2 
& 98.2 & 97.9 & 98.7 
& 97.1 & 96.4 & 96.8 
& 93.4 & 93.1 & 93.9 
& 92.8 & 92.4 & 93.2 
& 92.3 & 91.7 & 92.9 
& 95.9 & 95.3 & 95.9 
& 92.1 & 91.6 & 92.5 
& 56.6 & 2.97  \\
\multirow{1}{*}{M$^3$Att} & Swin-B & 97.5 & 97.4 & 97.9 
& 98.8 & 98.7 & 98.8 
& 96.9 & 95.6 & 96.8 
& 94.0 & 93.6 & 94.2 
& 95.5 & 94.9 & 95.3 
& 92.2 & 91.5 & 92.3 
& 96.1 & 96.2 & 96.4 
& 93.2 & 92.7 & 93.6 
& 183 & 12.14 \\
\multirow{1}{*}{DRIFA-Net} & ResNet18 & 98.4 & 98.4 & 98.7 
& 99.7 & 99.5 & 99.5 
& 98.2 & 97.5 & 98.6 
& 96.4 & 96.3 & 96.7 
& 98.2 & 97.9 & 98.5 
& 95.6 & 95.5 & 95.9 
& 97.0 & 96.8 & 97.1 
& 95.2 & 95.1 & 95.4 
& 53.8 & 4.83 \\
\approachsep
\multirow{1}{*}{HyPCA-Net18} & ResNet18 & 
98.8 & 98.7 & $\underline{99.0} $
& $\underline{99.8}$ & $\underline{99.8}$ & $\underline{99.8}$ & 
98.9 & 98.0 & $\underline{99.2} $& 97.2 & 97.1 & 97.5 
 & 99.4 & 99.3 & 99.7 
 & 95.7 & 95.7 & 96.1 
 & 98.3 & 97.9 & 98.5 
 & 96.3 & 96.2 & 96.5 
 & $\underline{14.47}$ & 2.25 \\
\multirow{1}{*}{HyPCA-Net50} & ResNet50 
& $\underline{99.2}$ & $\underline{99.1}$ & $\textbf{99.3}$ & $\textbf{99.9}$ & $\textbf{99.9}$ & $\textbf{99.9}$ & 
$\underline{99.1}$ & $\underline{98.3}$ & $\textbf{99.4} $
& $\underline{97.5}$ & $\underline{97.3}$ & $\underline{97.9}$ &
$\underline{99.6}$ & $\underline{99.6}$ & $\underline{99.8}$ & 
96.7 & 96.4 & 97.2 &
98.5 & $\underline{98.1}$ & $\underline{98.7}$ & 96.9 & 96.6 & 97.2 & 
28.4 & 3.03  \\
\multirow{1}{*}{HyPCA-Net-IN} & Inception-v3 & 98.5 & 98.4 & 98.6 & $\underline{99.8}$ & $\underline{99.8}$ & $\underline{99.8}$ & 
 98.4 & 97.7 & 98.9 & 96.9 & 96.9 & 97.2 & $\textbf{100}$ &$ \textbf{100}$ & $\textbf{100}$ 
 & $\underline{97.2}$ & $\underline{97.2}$ & $\underline{97.5}$ 
 & $\underline{98.7}$ & $\textbf{98.7}$ & $\underline{98.7} $
 & $\underline{97.1}$ & $\underline{97.1}$ & $\underline{97.3}$ 
 & 26.8 & 2.76 \\
 \multirow{1}{*}{HyPCA-Net-ViT} & ViT-Ti & $\textbf{99.4}$ & $\textbf{99.4}$ & $\textbf{99.3}$ & $\textbf{99.9}$ & $ \textbf{99.9}$ & $\textbf{99.9}$ & 
$ \textbf{99.2}$ & $\textbf{98.8}$ & $\underline{99.2}$ 
 & $\textbf{97.9}$ & $\textbf{97.7}$ & $\textbf{98.1}$ 
 & $\textbf{100}$ & $\textbf{100}$ & $\textbf{100} $
 & $\textbf{97.7}$ & $\textbf{97.7}$ & $\textbf{97.7} $
 & $\textbf{98.8}$ & $\textbf{98.7}$ & $\textbf{98.8} $
 & $\textbf{97.8}$ & $\textbf{97.5}$ & $\textbf{98.2}$ 
 & 22.5 & 3.42 \\
\hline
\end{tabular}
}
\vspace{-0.3cm}
\end{table*}


\begin{table}[htbp]
\centering
\footnotesize
\caption{Performance comparison of \texttt{HyPCA-Net-Seg} ($\texttt{SegNet}$ backbone) and $\texttt{HyPCA-Net-EMCAD}$ ($\texttt{EMCAD}$ backbone) with SOTA methods on $\texttt{D9}$–$\texttt{D10}$ datasets}
\vspace{-0.35cm}
\label{tab:tab2}
\scalebox{0.8}{
\setlength{\tabcolsep}{8pt}%
\renewcommand{\arraystretch}{1.1}%
\setlength{\doublerulesep}{0.0ex}
\newcommand{\approachsep}{\noalign{\vskip 0.1ex}\hline\hline\noalign{\vskip 0.1ex}}
\begin{tabular}{@{}l|cc|cc|cc@{}}
\hline
\multirow{2}{*}{Model} & \multicolumn{2}{c|}{D9: COVID-19} & \multicolumn{2}{c|}{D10: ISIC} & \multirow{2}{*}{$\#$P} & \multirow{2}{*}{$\#$F} 
\\ 
\cline{2-5}
 & DSC & IoU & DSC & IoU &  &  
\\ 
\hline
UNet & 47.7 & 38.6 & 87.3 & 80.2  & 34.5 & 65.5 
\\
PolypPVT & 81.7 & 74.4 & 90.4 & 83.9 
& 25.1 & \textbf{5.3} 
\\
MTTU-Net & 82.2 & 75.7 & 89.2 & 82.6 
& 71.6 & 20.9 
\\
MADGNet & 83.9 & 76.9 & 90.2 & 83.8 
& 31 & 14.2 
\\
PVT-CASCADE & 84.1 & 77.3 & 90.4 & 84.0 
& 34.1 & 7.6 
\\
EMCAD & 85.8 &  78.6 &  90.9 &  84.1 &  26.8 & \underline{5.6} 
\\
DRIFA-Net & 84.4 & 77.1 & 90.6 & 83.9 
& 67.3 & 19.9 
\\
\approachsep
HyPCA-Net-Seg & \underline{88.2} & \underline{80.9} & \underline{92.7} & \underline{85.3} 
& \underline{21.7} & 8.04 
\\
HyPCA-Net-EMCAD & \textbf{90.3} & \textbf{82.5} & \textbf{93.8} & \textbf{86.4} 
& \textbf{18.6} & 7.65 
\\
\hline
\end{tabular}}
\vspace{-0.5cm}
\end{table}



\noindent\textbf{(B) Multi-scale Multi-frequency Mutual Update Attention}
An \texttt{(MMMUA)} block (Figure \ref{fig:fig5} (A)) jointly learn multimodal information from inputs \( X' \) (via \texttt{Hy-SFA}) through mutual cross-modal interactions. Operating at multiple spatial scales and frequency bands, \texttt{MMMUA} generates multimodal spatial–frequency attention maps \(A_{D_i}\) that recalibrate each modality features, learning robust shared representations $X^s$ as:

\vspace{-0.2cm}
\begin{footnotesize}
\begin{equation} \label{eq:eq6}
X^s = X' \odot A_{D_i} \odot \omega_{i}.
\vspace{-.0cm}
\end{equation}
\end{footnotesize}
where channel-wise learnable parameters \( \omega_{i} \) for each modality $i$. To learn multimodal spatial–frequency attention maps \(A_{D_i}\), the \texttt{MMMUA} module comprises three core components: 

\begin{itemize}
\item \emph{ Frequency-domain channel Information Fusion} (\texttt{FCIF}) component (Figure \ref{fig:fig5}(C)) captures diverse channel dependencies by decomposing modality-specific input (e.g., \( x^{\prime}_{1} \)) into wavelet sub-bands via wavelet \texttt{DWC} denoted as \texttt{WDWC}: low-low (\texttt{LL}), high-low (\texttt{HL}), low-high (\texttt{LH}), and high-high (\texttt{HH}) \footnote{These sub-bands capture coarse-grained structures, horizontal edges, vertical edges, and fine‐grained textures, respectively.}.  
To learn diverse channel cues from each sub-band, we design a Hierarchical Channel Fusion (\texttt{HCF}) scheme (ref. \textcolor{blue}{Fig. 1 in the Supplementary}) that applies diverse global pooling operations -- \texttt{GAP}, \texttt{GMP}, \texttt{GMN}, and \texttt{GSP} -- to capture distinct channel contexts i.e., $C_{\texttt{GAP}}$, $C_{\texttt{GMP}}$, $C_{\texttt{GMN}}$, and $C_{\texttt{GSP}}$.  
These contexts are fused via inter-sub-band fusion (Eqs.\ref{eq:eq17}--\ref{eq:eq19}), integrating parallel and cascaded addition-with-subtraction to learn comprehensive and complementary frequency-domain channel details $C_f$: 


\vspace{-0.2cm}




\settowidth{\lhsw}{$(\texttt{LL},\,\texttt{HL},\,\texttt{LH},\,\texttt{HH})$} 

\begin{footnotesize}
\begin{equation}\label{eq:eq16}
\begin{alignedat}{3}
&\mathmakebox[\lhsw][l]{(\mathrm{LL},\,\mathrm{HL},\,\mathrm{LH},\,\mathrm{HH})} &{}={}&
  \mathrm{WDWC}(x'_1),\\
&\mathmakebox[\lhsw][l]{[C_{\text{Sum}},\,C_{\text{Diff}}]} &{}={}&
  \mathrm{HCF}(\mathrm{LL},\,\mathrm{HL},\,\mathrm{LH},\,\mathrm{HH}),\\
&\mathmakebox[\lhsw][l]{C_f} &{}={}&
  \sum_{j} \omega_j \odot \mathrm{DR}_j\!\bigl(\mathrm{BN}_j(C_j)\bigr)
\end{alignedat}
\end{equation}
\end{footnotesize}
where $C_j = [C_{\text{Sum}},\,C_{\text{Diff}}]$ and \texttt{DR} and \texttt{BN} denote dropout and batch norm. \texttt{HCF} can be written as:



\begin{footnotesize}
\begin{equation} \label{eq:eq17}
\begin{aligned}
C_{\mathrm{GAP}} &= \sum_{T} \mathrm{GAP}(T), \quad
C_{\mathrm{GMP}} = \sum_{T} \mathrm{GMP}(T), \\
C_{\mathrm{GMN}} &= \sum_{T} \mathrm{GMN}(T), \quad
C_{\mathrm{GSP}} = \sum_{T} \mathrm{GSP}(T) \\
\end{aligned}
\end{equation}
\end{footnotesize}
where $T \in \{\mathrm{LL}, \mathrm{LH}, \mathrm{HL}, \mathrm{HH} \}$. 

\settowidth{\lhsw}{$C_{\texttt{Diff}}$} 

\begin{footnotesize}
\begin{equation}\label{eq:eq18}
\begin{alignedat}{3}
&\mathmakebox[\lhsw][l]{C_{\text{Sum}}}  &{}={}&
  C_{\mathrm{GAP}} + C_{\mathrm{GMP}} + C_{\mathrm{GMN}} + C_{\mathrm{GSP}},\\
&\mathmakebox[\lhsw][l]{C_{\text{Diff}}} &{}={}&
  C_{\mathrm{GMP}} - \bigl(C_{\mathrm{GAP}} + C_{\mathrm{GMN}}\bigr)
\end{alignedat}
\end{equation}
\end{footnotesize}

\item  \emph{Spatial-domain Multi-scale Information Fusion} (\texttt{SMIF}) (Figure~\ref{fig:fig5} (D)) captures fine-grained multi-scale spatial-channel 
Specifically, it stacks an \texttt{MSHC} block followed by our \texttt{SCPFA} block from the \texttt{SCALA} module to refine representations ($\hat{x}'_{m}$) across scales. Next, diverse global pooling operations -- \texttt{GAP}, \texttt{GMP}, \texttt{GMN}, and \texttt{GSP} -- are applied to capture complementary channel contexts, which are fused to learn diverse spatial–channel dependencies i.e., $C_{\texttt{SP}}$ as: 

\vspace{-0.3cm}



\settowidth{\lhsw}{$C_{SP}$} 


\begin{footnotesize}
\begin{equation}\label{eq:eq19}
\begin{alignedat}{3}
&\mathmakebox[\lhsw][l]{\hat{x}'_{m}} &{}={}& \mathrm{SCPFA}\bigl(\mathrm{MSHC}(x'_{m})\bigr),\\
&\mathmakebox[\lhsw][l]{C_{\texttt{SP}}}   &{}={}& \mathrm{GMP}(\hat{x}'_{m})
  + \mathrm{GAP}(\hat{x}'_{m})
  + \mathrm{GMN}(\hat{x}'_{m})
  + \mathrm{GSP}(\hat{x}'_{m}).
\end{alignedat}
\end{equation}
\end{footnotesize}



\item  \emph{Mutual Cross Bidirectional Interactions} (\texttt{MCBI}) component (Figure \ref{fig:fig5}(B)) is designed to integrate frequency domain channel contexts -- i.e., $C_f$ from \texttt{FCIF} with spatial domain spatial-channel details -- i.e., $C_{\texttt{SP}}$ from \texttt{SMIF} for multimodal input \(X^{\prime}\), facilitating \emph{mutual enhancement through complementary information exchange}. It applies \emph{bidirectional cross-interactions} to fuse $C_f$ of modality \(x^{\prime}_{1}\) with $C_{\texttt{SP}}$ of modality \(x^{\prime}_{m}\), and vice versa.
This asymmetric interaction mutually updates channel dependencies while preserving modality-specific features, capturing complementary cross-domain spatial-channel information. An adaptive scaling factor \(w_i\) modulates these interactions to enhance discriminative capacity. 
Finally, sigmoid activations generate cross-domain attention maps: \(\mathcal{A}_{D_i} = \texttt{MMMUA}(\cdot)\), which highlight discriminative regions and jointly optimize spatial-channel coherence across modalities. The attention map is computed as:

\vspace{-0.2cm}

\settowidth{\lhsw}{$A_{D_i}$}

\begin{footnotesize}
\begin{equation}\label{eq:eq20}
\begin{alignedat}{3}
&\mathmakebox[\lhsw][l]{\hat{C}_{\texttt{SP}}} &{}={}&
  \omega_{1}\odot C_{\texttt{SP}} + \omega_{2}\odot C_{f},\\
&\mathmakebox[\lhsw][l]{\hat{C}_{f}}  &{}={}&
  \bigl(\mathrm{DR}(\mathrm{BN}(\hat{C}_{SP}))\odot \omega_{3}\bigr)
  + \bigl(C_{f}\odot \omega_{2}\bigr),\\
&\mathmakebox[\lhsw][l]{A_{D_i}}      &{}={}&
  \bigl[\,\sigma_{1}(\hat{C}_{f}),\ \sigma_{2}(\hat{C}_{\texttt{SP}})\,\bigr].
\end{alignedat}
\end{equation}
\end{footnotesize}

\end{itemize}


\noindent\textbf{Summary.} \texttt{HyPCA} couples parallel spatial–channel fusion in \texttt{RALA} 
(\texttt{SCPFA} = \texttt{CHIA} $\parallel$ \texttt{SHIA}) with cascaded hybrid-space, dual-domain modeling in DVCA 
(\texttt{Hy-SFA} = \texttt{TFSI} $\to$ \texttt{FDCA}; 
\texttt{MMMUA} = \texttt{FCIF} + \texttt{SMIF} + \texttt{MCBI}). 
The parallel step curbs early information loss, while the cascade fuses token- and frequency-space evidence 
and synchronizes spatial and frequency cues across modalities (ref. Figs.~\ref{fig:fig2}--\ref{fig:fig5}). 
Such parallel $\to$ cascaded coupling is new to multimodal fusion in medical imaging and yields higher 
performance gains at low computational cost (Tables~\ref{tab:tab1}--\ref{tab:tab6}).


\begin{table*}[htbp]
\centering
\footnotesize
\caption{\small Ablation of \texttt{HyPCA-Net18} modules (left) and components (right), highlighting their contributions to performance gains on \texttt{D5}–\texttt{D6} datasets.}
\vspace{-0.3cm}
\label{tab:tab3}
\scalebox{0.725}{
\setlength{\tabcolsep}{5.4pt}%
\renewcommand{\arraystretch}{1.1}%
\setlength{\doublerulesep}{0.95ex}
\newcommand{\approachsep}{\noalign{\vskip 0.1ex}\hline\hline\noalign{\vskip 0.1ex}}
\begin{tabular}{@{}ccc|c|c|c||ccc|ccc|cc|cc|cc|cc@{}}
\hline
\multicolumn{3}{c|}{\textbf{Integrated modules}} &
\multicolumn{1}{c|}{\textbf{D5: HAM10000}}
 & \multicolumn{1}{c|}{\textbf{D6: SIPaKMeD}}
 & \multicolumn{1}{c||}{}
 & \multicolumn{8}{c|}{\textbf{Integrated components within module}}
 & \multicolumn{2}{c|}{\textbf{D5: HAM10000}}
 & \multicolumn{2}{c|}{\textbf{D6: SIPaKMeD}}
 & \multicolumn{2}{c}{} \\
\hline
RALA & Hy-SFA & MMMUA & F1 & F1 & $\#$P
 & MSHC & CHIA & SHIA & FCIF & SMIF & MCBI & TFSI & FDCA & Acc & F1
 & Acc & F1
 & $\#$P & $\#$F \\
\hline
$\times$ & $\times$ & $\times$ & 93.2 
& 89.3 & 23.4 &
$\times$ & $\times$ & $\times$ & $\times$ & $\times$ & $\times$ & $\times$ & $\times$ & 93.2 & 93.1 
& 89.5 & 89.3 & 23.4 & \textbf{1.16} \\
$\times$ & $\times$ & $\checkmark$ & 97.1 
& 93.0 & 25.2 &
$\checkmark$ & $\times$ & $\times$ & $\checkmark$ & $\times$ & $\times$ & $\checkmark$ & $\times$ & 96.8 & 96.6 
& 92.9 & 92.8 & \textbf{11.1} & \underline{1.8} \\
$\times$ & $\checkmark$ & $\times$ & 96.7 
& 92.8 & 28.1 
& $\checkmark$ & $\checkmark$ & $\times$ & $\checkmark$ & $\times$ & $\times$ & $\checkmark$ & $\times$ & 97.6 & 97.6 
& 93.9 & 93.7 & 12.4 & 1.9 \\
$\times$ & $\checkmark$ & $\checkmark$ & \underline{98.6} 
& \underline{94.9} & 30.5 & 
 $\times$ & $\times$ & $\checkmark$ & $\checkmark$ & $\times$ & $\times$ & $\checkmark$ & $\times$ & 97.3 & 97.2 
 & 93.6 & 93.4 & 28.6 & 2.84 \\
\checkmark & $\times$ & $\times$ & 96.1 
& 92.3 & \textbf{10.1} &
 $\checkmark$ & $\checkmark$ & $\checkmark$ & $\checkmark$ & $\times$ & $\times$ & $\checkmark$ & $\times$ & 97.9 & 97.7 
 & 94.3 & 94.2 & 12.7 & 1.92 \\
$\checkmark$  & $\times$ & $\checkmark$  & 97.8 
& 93.8 & \underline{11.9} &  $\checkmark$ & $\times$ & $\times$ & $\checkmark$ & $\checkmark$ & $\checkmark$ & $\checkmark$ & $\times$ & 98.3 & 98.3 
& 94.3 & 94.0 & \underline{11.7} & 1.85 \\
$\checkmark$  & $\checkmark$  & $\times$ & 97.3  
& 93.4  & 12.9 &
 $\checkmark$ & $\checkmark$ & $\checkmark$ & $\checkmark$ & $\checkmark$ & $\checkmark$ & $\checkmark$ & $\times$ & \underline{98.7} & \underline{98.7} 
 & \underline{95.3} & \underline{95.1} & 13.4 & 1.91 \\
$\checkmark$ & $\checkmark$ & $\checkmark$ & \textbf{99.3}  
& \textbf{95.7}  & 14.5 &
$\checkmark$ & $\checkmark$ & $\checkmark$ & $\checkmark$ & $\checkmark$ & $\checkmark$ & $\checkmark$ & $\checkmark$ & \textbf{99.4} & \textbf{99.3} 
& \textbf{95.7} 
& \textbf{95.7} & 14.5 & 2.25 \\
 \hline
\end{tabular}
}
\vspace{-0.3cm}
\end{table*}


\begin{table*}[htbp]
\centering
\footnotesize
\caption{\small Ablation of \texttt{HyPCA-Net-EMCAD} modules (left) and components (right), highlighting their contributions to performance gains on \texttt{D9}–\texttt{D10} datasets.}
\vspace{-0.3cm}
\label{tab:tab4}
\scalebox{0.725}{
\setlength{\tabcolsep}{6.2pt}%
\renewcommand{\arraystretch}{1.1}%
\setlength{\doublerulesep}{0.95ex}
\newcommand{\approachsep}{\noalign{\vskip 0.1ex}\hline\hline\noalign{\vskip 0.1ex}}

\begin{tabular}{@{}ccc|cc|cc|c||ccc|ccc|cc|cc|cc|cc@{}}
\hline
\multicolumn{3}{c|}{\textbf{Integrated modules}} &
\multicolumn{2}{c|}{\textbf{D9: COVID-19}}
 & \multicolumn{2}{c|}{\textbf{D10: ISIC}}
 & \multicolumn{1}{c||}{}
 & \multicolumn{8}{c|}{\textbf{Integrated components within module}}
 & \multicolumn{2}{c|}{\textbf{D9: COVID-19}}
 & \multicolumn{2}{c|}{\textbf{D10: ISIC}}
 & \multicolumn{2}{c}{} \\
\hline
RALA & Hy-SFA & MMMUA & DSC & IoU & DSC & IoU & \#P
 & MSHC & CHIA & SHIA & FCIF & SMIF & MCBI & TFSI & FDCA & DSC & IoU & DSC & IoU
 & $\#$P & $\#$F \\
\hline
$\times$ & $\times$ & $\times$ & 82.8 & 75.5 
& 85.9 & 79.1 & 27.2 &
$\times$ & $\times$ & $\times$ & $\times$ & $\times$ & $\times$ & $\times$ & $\times$ & 82.8 & 75.5 
& 85.9 & 79.1 & 27.2 & \textbf{4.32} \\

$\times$ & $\times$ & $\checkmark$  & 86.2 & 78.8 
& 89.1 & 81.9 & 29.2 &
$\checkmark$ & $\times$ & $\times$ & $\checkmark$ & $\times$ & $\times$ & $\checkmark$ & $\times$ & 85.5 & 78.2 
& 87.4 & 81.2 & \textbf{15.1} & \underline{6.27} \\
$\times$ & $\checkmark$  & $\times$ & 85.5 & 77.9 
& 88.4 & 81.1 & 32.1 &
$\checkmark$ & $\checkmark$ & $\times$ & $\checkmark$ & $\times$ & $\times$ & $\checkmark$ & $\times$ & 86.5 & 79.1 
& 88.9 & 82.5 & 16.2 & 6.58 \\
$\times$ & $\checkmark$  & $\checkmark$ & \underline{88.5} & \underline{80.9} 
& \underline{90.7} & \underline{84.1} & 34.4 & 
 $\times$ & $\times$ & $\checkmark$ & $\checkmark$ & $\times$ & $\times$ & $\checkmark$ & $\times$ & 86.2 & 78.8 
 & 88.5 & 82.1 & 32.4 & 9.46 \\
$\checkmark$  & $\times$ & $\times$ & 84.7 & 77.4 
& 87.9 & 80.8 & \textbf{13.9} 
& $\checkmark$ & $\checkmark$ & $\checkmark$ & $\checkmark$ & $\times$ & $\times$ & $\checkmark$ & $\times$ & 86.8 & 79.5 
& 89.3 & 82.8 & 16.5 & 6.64 \\
$\checkmark$  & $\times$ & $\checkmark$  & 88.1 & 80.6 
& 91.2 & 83.6 & \underline{15.7} &  $\checkmark$ & $\times$ & $\times$ & $\checkmark$ & $\checkmark$ & $\checkmark$ & $\checkmark$ & $\times$ & 87.6 & 80.1 
& 90.6 & 83.7 & \underline{15.5} & 6.43 \\
$\checkmark$  & $\checkmark$  & $\times$ & 87.4 & 79.7  
& 90.3 & 82.8  & 16.6 &
 $\checkmark$ & $\checkmark$ & $\checkmark$ & $\checkmark$ & $\checkmark$ & $\checkmark$ & $\checkmark$ & $\times$ & \underline{89.1} & \underline{81.0} 
 & \underline{92.7} & \underline{85.5} & 17.2 & 6.61 \\
$\checkmark$ & $\checkmark$ & $\checkmark$ & \textbf{90.3} & \textbf{82.5}  
& \textbf{93.8} & \textbf{86.4}  & 18.6 &
$\checkmark$ & $\checkmark$ & $\checkmark$ & $\checkmark$ & $\checkmark$ & $\checkmark$ & $\checkmark$ & $\checkmark$ & \textbf{90.3} & \textbf{82.5} 
& \textbf{93.8} & \textbf{86.4} & 18.6 & 7.65 \\
\hline
\end{tabular}
}
\vspace{-0.3cm}
\end{table*}






\begin{figure*}[ht!]
    \centering
    \includegraphics[width=0.8\textwidth]{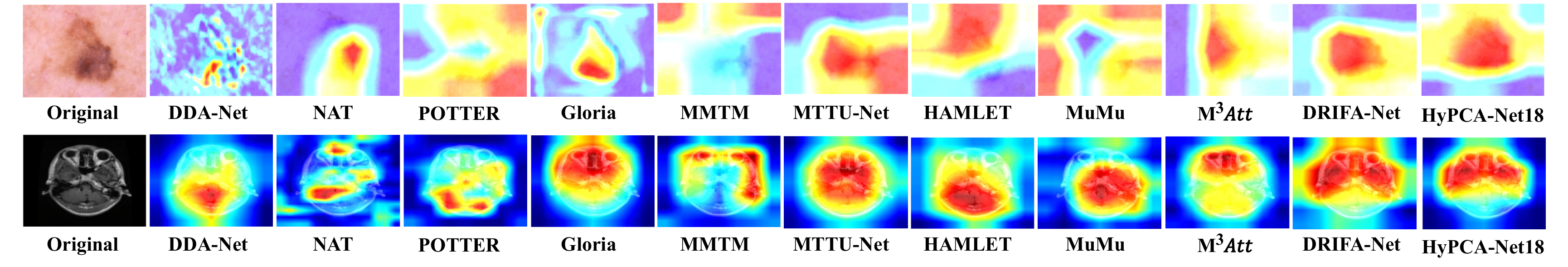}
    \vspace{-0.3cm}
    \caption{
    \small Visual representation of the important regions highlighted by our proposed $\texttt{HyPCA-Net}$ framework and ten other SOTA methods using the
$\texttt{GRAD-CAM}$ technique on two benchmark datasets $\texttt{D5}$ and $\texttt{D1}$.
    }
    \label{fig:fig6}
    \vspace{-0.4cm}
\end{figure*}



\section{Experimental Analysis and Results}
\vspace{-0.2cm}
\textbf{Datasets} We evaluated \texttt{HyPCA-Net} on ten public medical‐imaging benchmarks:
For classification we used Nickparvar \cite{nickparvar2021}, $\texttt{IQ-OTH NCCD}$ \cite{alyasriy2020iq}, $\texttt{Tuberculosis}$ \cite{rahman2020reliable}, $\texttt{CNMC-2019}$ \cite{mourya2019all}, $\texttt{HAM10000}$ \cite{tschandl2018ham10000}, $\texttt{SIPaKMeD}$ \cite{plissiti2018sipakmed}, $\texttt{CRC}$ \cite{kather2016multi}, $\texttt{CBIS-DDSM}$ \cite{sawyer2016curated}, denoted as ($\texttt{D1}$–$\texttt{D8}$). For segmentation we used \texttt{COVID-19} lung $\texttt{CT}$ \cite{jun2020covid} and $\texttt{ISIC2018}$ skin lesions \cite{codella2019skin} datasets, denoted as ($\texttt{D9}$–$\texttt{D10}$). 
Images were resized to $128\times128\times3$ (classification) or $224\times224\times3$ (segmentation) and a split of $80/10/10$ (train/val/test) with standard on-the-fly augmentations was used.

\noindent
\textbf{Models} We evaluated $\texttt{HyPCA-Net}$ against a comprehensive set of state-of-the-art (SOTA) baselines. For classification, unimodal baselines are $\texttt{DDA-Net}$, $\texttt{MADGNet}$’s $\texttt{MFMSA}$ \cite{nam2024modality}, and \texttt{EMCAD}'s $\texttt{MSCAM}$ \cite{rahman2024emcad} -- denoted as $\texttt{M1}$–$\texttt{M3}$; while multimodal fusion baselines comprises $\texttt{Gloria}$ \cite{huang2021gloria}, $\texttt{MTTU-Net}$ \cite{cheng2022fully}, $\texttt{MuMu}$ \cite{islam2022mumu}, $\texttt{M$^3$3Att}$, and $\texttt{DRIFA-Net}$ -- denoted as $\texttt{M4}$–$\texttt{M8}$.  
For segmentation, we compare $\texttt{MTTU-Net}$ and $\texttt{DRIFA-Net}$ (with $\texttt{SegNet}$ decoder), alongside SOTA unimodal baselines: $\texttt{UNet}$ \cite{ronneberger2015u}, 
$\texttt{PolypPVT}$ \cite{dong2021polyp}, 
$\texttt{MADGNet}$, $\texttt{EMCAD}$, and
\texttt{PVT-CASCADE} \cite{rahman2023medical}, 
denoted as $\texttt{M9}$–$\texttt{M15}$. 
Notably, we instantiate \texttt{HyPCA-Net} with four classification backbones -- $\texttt{ResNet-18}$ \cite{he2016deep}, $\texttt{ResNet-50}$ \cite{he2016deep}, $\texttt{Inception-v3}$ \cite{szegedy2016rethinking}, and $\texttt{ViT-Ti}$ \cite{steiner2021train}, and $\texttt{SegNet}$ \cite{badrinarayanan2017segnet} and $\texttt{EMCAD}$ \cite{rahman2024emcad} decoders (paired with our $\texttt{HyPCA-Net18}$ encoder) for segmentation. These variants are denoted as $\texttt{HyPCA-Net18},$ $\texttt{HyPCA-Net50}$, $\texttt{HyPCA-Net-IN}$, $\texttt{HyPCA-Net-ViT}$, $\texttt{HyPCA-Net-Seg}$, and $\texttt{HyPCA-Net-EMCAD}$.

\noindent\textbf{Notation} We report Accuracy ($\texttt{Acc}$), F1 score (\texttt{F1}), $\texttt{AUC}$, Dice Score ($\texttt{DSC}$), Intersection-over-Union ($\texttt{IoU}$), number of parameters in millions ($\#$P), and GFLOPs ($\#$F) as metrics.

\noindent
\emph{See \textcolor{blue}{supplementary} for dataset, model, and training details.}


\begin{table}[t]
\centering
\begin{footnotesize}
\caption{\small Performance comparison (F1-score) of cascaded (\texttt{CA}) vs. hybrid ($\texttt{HA}$) attention in $\texttt{CHIA}$ and $\texttt{SHIA}$ on \texttt{D5} and $\texttt{D6}$, keeping other components fixed.}
\vspace{-0.2cm}
\label{tab:tab6}
\setlength{\tabcolsep}{6pt}
\scalebox{0.9}{
\begin{tabular}{cccccc}
\hline
\textbf{HA} & \textbf{CA} & \textbf{D5} & \textbf{D6} & \textbf{$\#$P} & \textbf{$\#$F} \\
\hline
$\checkmark$ & $\times$ & \textbf{99.3} & \textbf{95.7} & 14.47 & 2.25  \\
$\times$ & $\checkmark$ & 98.9 & 95.2 & 14.47 & 2.25 \\
\hline
\end{tabular}
}
\end{footnotesize}
\vspace{-0.5cm}
\end{table}


\vspace{-0.1cm}
\subsection{Performance Comparisons}
\vspace{-0.1cm}

We present results in Tables~\ref{tab:tab1} and~\ref{tab:tab2}, from which it can be seen that \texttt{HyPCA-Net} achieves exceptional performance on classification and segmentation datasets, with performance ranging from $80.9\%$ to $100\%$. 
Compared to SOTA unimodal and multimodal fusion baselines, \texttt{HyPCA-Net} achieves performance improvements of $0.1\%–43.9\%$, while reducing parameters by up to $92\%$ and \texttt{FLOPs} by up to $81.47\%$.
We conducted a qualitative analysis on  \texttt{D1} and \texttt{D5} datasets as shown in Figure \ref{fig:fig6}, which further validate its effectiveness. \emph{See \textcolor{blue}{supplementary} for more details and results}.

As we discussed earlier, to address \textbf{Challenge 1}, our \texttt{HyPCA-Net} achieves an effective balance between optimal performance and minimal computational cost by incorporating two efficient and effective modules -- \texttt{RALA} and \texttt{DVCA}. 
Because cascaded designs are inherently \textit{sequential}, they process different aspects in isolation, lack joint optimization, and are therefore \textit{limited in preserving holistic information}, ultimately \textit{constraining the richness of learned representations} \cite{lv2024lightweight1, shen2021parallel}. In contrast, our hybrid attention design seamlessly integrates parallel fusion attention with cascaded attention to preserve holistic cues while \emph{jointly} and \emph{cascadingly} modeling spatial–channel dependencies in both hybrid space and dual-domain representation learning. As a result, \texttt{HyPCA-Net} overcomes the intrinsic drawbacks of purely cascaded schemes and directly addresses \textbf{Challenge 2}, yielding more robust shared representation learning. \emph{See discussion part of \textcolor{blue}{supplementary} material}.




\vspace{-0.1cm}
\subsection{Ablation Study}\label{ablation}
\vspace{-0.1cm}
We evaluated key components of \texttt{HyPCA-Net} on datasets \texttt{D5}-\texttt{D6} and \texttt{D9}-\texttt{D10}, focusing on the \texttt{RALA} (with \texttt{SCPFA}), and \texttt{Hy-SFA} and \texttt{MMMUA} modules within \texttt{DVCA}. Tables~\ref{tab:tab3}-\ref{tab:tab4} show that \texttt{HyPCA-Net} outperforms its ablated variants by $0.4\%-7.9\%$, highlighting the effectiveness of its hybrid attention design. 
Further, we analyzed contributions from individual modules: \texttt{MSHA}, \texttt{CHIA}, and \texttt{SHIA} in \texttt{SCALA}; \texttt{FCIF}, \texttt{SMIF}, and \texttt{MCBI} in \texttt{MMMUA}; and \texttt{TFSI} and \texttt{FDCA} in \texttt{Hy-SFA} (Tables~\ref{tab:tab3}-\ref{tab:tab4}). \texttt{HyPCA-Net} achieves up to $7.9\%$ performance improvements over partial configurations, emphasizing the complementary roles of these modules. Finally, we compare cascaded (\texttt{CHIA}~$\to$~\texttt{SHIA}~$\to$~\texttt{DVCA}) and hybrid (parallel \texttt{CHIA}~+~\texttt{SHIA}~$\to$~\texttt{DVCA}) configurations (Table~\ref{tab:tab6}); The hybrid variant achieves a $0.4\%-0.5\%$ performance gain, capturing more robust shared representations at lower computational cost.

One can infer that the limited performance of the ablated variants (Tables \ref{tab:tab3}–\ref{tab:tab4}) arises from their inability to learn robust shared representations at low computational cost-essential for \textbf{addressing challenge 1}. 
To \textbf{address challenge 2}, we note that cascaded attention architectures face progressive information loss due to the absence of \textit{parallel–cascaded (hybrid) fusion}, which hinders holistic information preservation and weakens representation learning, ultimately lowering performance (Table~\ref{tab:tab6})~\cite{shen2021parallel, lv2024lightweight1}. As discussed earlier, \texttt{HyPCA-Net} addresses both challenges through its integrated hybrid attention modules (\texttt{RALA}, \texttt{Hy-SFA}, and \texttt{MMMUA}). Specifically, \texttt{RALA} efficiently captures multi-scale spatial–channel dependencies via parallel fusion, while \texttt{Hy-SFA} and \texttt{MMMUA} model complementary dual-domain interactions. 
Our hybrid framework preserves holistic information, improving multi-disease classification with low cost (see \emph{\textcolor{blue}{supplementary}} for more results)

\vspace{-0.25cm}
\section{Conclusion} \label{sec:concl}
\vspace{-0.1cm}
We present \texttt{HyPCA-Net}, an efficient multimodal fusion framework designed for advancing medical image analysis, making it highly suitable for real-world medical AI applications. \texttt{HyPCA-Net} is evaluated on ten diverse datasets, demonstrating strong generalizability across medical imaging modalities. It outperforms leading baselines with up to $\approx 5.2\%$ higher performance and $\approx 73.1\%$ lower computational cost.
In the future, we are interested in integrating \emph{adversarial robustness}, and \emph{domain-shift adaptation} into \texttt{HyPCA-Net} for \emph{robust generalization}.

{
    \small
    \bibliographystyle{ieeenat_fullname}

\begin{thebibliography}{39}
\providecommand{\natexlab}[1]{#1}
\providecommand{\url}[1]{\texttt{#1}}
\expandafter\ifx\csname urlstyle\endcsname\relax
  \providecommand{\doi}[1]{doi: #1}\else
  \providecommand{\doi}{doi: \begingroup \urlstyle{rm}\Url}\fi

\bibitem[Alyasriy and Muayed(2020)]{alyasriy2020iq}
Hamdalla Alyasriy and A Muayed.
\newblock The iq-othnccd lung cancer dataset.
\newblock \emph{Mendeley Data}, 1\penalty0 (1):\penalty0 1--13, 2020.

\bibitem[Badrinarayanan et~al.(2017)Badrinarayanan, Kendall, and Cipolla]{badrinarayanan2017segnet}
Vijay Badrinarayanan, Alex Kendall, and Roberto Cipolla.
\newblock Segnet: A deep convolutional encoder-decoder architecture for image segmentation.
\newblock \emph{IEEE transactions on pattern analysis and machine intelligence}, 39\penalty0 (12):\penalty0 2481--2495, 2017.

\bibitem[Chakladar et~al.(2021)Chakladar, Kumar, Roy, Dogra, Scheme, and Chang]{chakladar2021multimodal}
Debashis~Das Chakladar, Pradeep Kumar, Partha~Pratim Roy, Debi~Prosad Dogra, Erik Scheme, and Victor Chang.
\newblock A multimodal-siamese neural network (msnn) for person verification using signatures and {EEG}.
\newblock \emph{Information Fusion}, 71:\penalty0 17--27, 2021.

\bibitem[Chen et~al.(2018)Chen, Rubanova, Bettencourt, and Duvenaud]{Chen2018NeuralODE}
Ricky T.~Q. Chen, Yulia Rubanova, Jesse Bettencourt, and David~K. Duvenaud.
\newblock Neural ordinary differential equations.
\newblock In \emph{Advances in Neural Information Processing Systems (NeurIPS)}, 2018.

\bibitem[Cheng et~al.(2022)Cheng, Liu, Kuang, and Wang]{cheng2022fully}
J. Cheng, J. Liu, H. Kuang, and J. Wang.
\newblock A fully automated multimodal mri-based multi-task learning for glioma segmentation and idh genotyping.
\newblock \emph{IEEE Transactions on Medical Imaging}, 41\penalty0 (6):\penalty0 1520--1532, 2022.

\bibitem[Codella et~al.(2019)Codella, Rotemberg, Tschandl, Celebi, Dusza, Gutman, Helba, Kalloo, Liopyris, Marchetti, et~al.]{codella2019skin}
Noel Codella, Veronica Rotemberg, Philipp Tschandl, M~Emre Celebi, Stephen Dusza, David Gutman, Brian Helba, Aadi Kalloo, Konstantinos Liopyris, Michael Marchetti, et~al.
\newblock Skin lesion analysis toward melanoma detection 2018: A challenge hosted by the international skin imaging collaboration (isic).
\newblock \emph{arXiv preprint arXiv:1902.03368}, 2019.

\bibitem[Cui et~al.(2023)Cui, Tao, Ren, and Knoll]{cui2023dual}
Y. Cui, Y. Tao, W. Ren, and A. Knoll.
\newblock Dual-domain attention for image deblurring.
\newblock In \emph{Proceedings of the AAAI Conference on Artificial Intelligence}, pages 479--487, 2023.

\bibitem[Das et~al.(2024)Das, Khare, Chakladar, and Jaluka]{das2024fusion}
Khakon Das, Ashish Khare, Debashis~Das Chakladar, and Dinesh Jaluka.
\newblock Fusion in medical imaging techniques for enhancing stroke region detection: A selective review.
\newblock In \emph{International Conference on Pattern Recognition}, pages 167--179. Springer, 2024.

\bibitem[Dhar et~al.(2024)Dhar, Rana, and Goyal]{dhar2024uncertainty}
Joy Dhar, Kapil Rana, and Puneet Goyal.
\newblock Uncertainty-rifa-net: Uncertainty aware robust information fusion attention network for brain tumors classification in mri images.
\newblock In \emph{International Conference on Pattern Recognition}, pages 311--327. Springer, 2024.

\bibitem[Dhar et~al.(2025)Dhar, Zaidi, Haghighat, Roy, Goyal, Alavi, and Kumar]{dhar2024multimodal}
Joy Dhar, Nayyar Zaidi, Maryam Haghighat, Sudipta Roy, Puneet Goyal, Azadeh Alavi, and Vikas Kumar.
\newblock Multimodal fusion learning with dual attention for medical imaging.
\newblock In \emph{2025 IEEE/CVF Winter Conference on Applications of Computer Vision (WACV)}, pages 4362--4371. IEEE, 2025.

\bibitem[Dong et~al.(2021)Dong, Wang, Fan, Li, Fu, and Shao]{dong2021polyp}
Bo Dong, Wenhai Wang, Deng-Ping Fan, Jinpeng Li, Huazhu Fu, and Ling Shao.
\newblock Polyp-pvt: Polyp segmentation with pyramid vision transformers.
\newblock \emph{arXiv preprint arXiv:2108.06932}, 2021.

\bibitem[Hassani et~al.(2023)Hassani, Walton, Li, Li, and Shi]{hassani2023neighborhood}
Ali Hassani, Steven Walton, Jiachen Li, Shen Li, and Humphrey Shi.
\newblock Neighborhood attention transformer.
\newblock In \emph{Proceedings of the IEEE/CVF Conference on Computer Vision and Pattern Recognition}, pages 6185--6194, 2023.

\bibitem[He et~al.(2016)He, Zhang, Ren, and Sun]{he2016deep}
Kaiming He, Xiangyu Zhang, Shaoqing Ren, and Jian Sun.
\newblock Deep residual learning for image recognition.
\newblock In \emph{Proceedings of the IEEE conference on computer vision and pattern recognition}, pages 770--778, 2016.

\bibitem[He et~al.(2023)He, Wang, Zhao, and Chen]{He2023}
X. He, Y. Wang, S. Zhao, and X. Chen.
\newblock Co-attention fusion network for multimodal skin cancer diagnosis.
\newblock \emph{Pattern Recognition}, 133:\penalty0 108990, 2023.

\bibitem[Huang et~al.(2021)Huang, Shen, Lungren, and Yeung]{huang2021gloria}
S.~C. Huang, L. Shen, M.~P. Lungren, and S. Yeung.
\newblock Gloria: A multimodal global-local representation learning framework for label-efficient medical image recognition.
\newblock In \emph{Proceedings of the IEEE/CVF International Conference on Computer Vision}, pages 3942--3951, 2021.

\bibitem[Islam and Iqbal(2020)]{islam2020hamlet}
Md~Mofijul Islam and Tariq Iqbal.
\newblock Hamlet: A hierarchical multimodal attention-based human activity recognition algorithm.
\newblock In \emph{2020 IEEE/RSJ International Conference on Intelligent Robots and Systems (IROS)}, pages 10285--10292. IEEE, 2020.

\bibitem[Islam and Iqbal(2022)]{islam2022mumu}
M.~M. Islam and T. Iqbal.
\newblock Mumu: Cooperative multitask learning-based guided multimodal fusion.
\newblock In \emph{Proceedings of the AAAI Conference on Artificial Intelligence}, pages 1043--1051, 2022.

\bibitem[Joze et~al.(2020)Joze, Shaban, Iuzzolino, and Koishida]{joze2020mmtm}
Hamid Reza~Vaezi Joze, Amirreza Shaban, Michael~L Iuzzolino, and Kazuhito Koishida.
\newblock Mmtm: Multimodal transfer module for cnn fusion.
\newblock In \emph{Proceedings of the IEEE/CVF conference on computer vision and pattern recognition}, pages 13289--13299, 2020.

\bibitem[Jun et~al.(2020)Jun, Cheng, Yixin, Xingle, Jiantao, Ziqi, Minqing, Xin, Xueyuan, Shucheng, et~al.]{jun2020covid}
Ma Jun, Ge Cheng, Wang Yixin, An Xingle, Gao Jiantao, Yu Ziqi, Zhang Minqing, Liu Xin, Deng Xueyuan, Cao Shucheng, et~al.
\newblock Covid-19 ct lung and infection segmentation dataset.
\newblock \emph{(No Title)}, 2020.

\bibitem[Kather et~al.(2016)Kather, Weis, Bianconi, Melchers, Schad, Gaiser, Marx, and Z{\"o}llner]{kather2016multi}
Jakob~Nikolas Kather, Cleo-Aron Weis, Francesco Bianconi, Susanne~M Melchers, Lothar~R Schad, Timo Gaiser, Alexander Marx, and Frank~Gerrit Z{\"o}llner.
\newblock Multi-class texture analysis in colorectal cancer histology.
\newblock \emph{Scientific reports}, 6\penalty0 (1):\penalty0 1--11, 2016.

\bibitem[Liu et~al.(2023)Liu, Ding, Zhang, and Jiang]{liu2023multi}
Chang Liu, Henghui Ding, Yulun Zhang, and Xudong Jiang.
\newblock Multi-modal mutual attention and iterative interaction for referring image segmentation.
\newblock \emph{IEEE Transactions on Image Processing}, 32:\penalty0 3054--3065, 2023.

\bibitem[Liu et~al.(2021)Liu, Lin, Cao, Hu, Wei, Zhang, Lin, and Guo]{Liu_2021_ICCV}
Ze Liu, Yutong Lin, Yue Cao, Han Hu, Yixuan Wei, Zheng Zhang, Stephen Lin, and Baining Guo.
\newblock Swin transformer: Hierarchical vision transformer using shifted windows.
\newblock In \emph{Proceedings of the IEEE/CVF International Conference on Computer Vision (ICCV)}, pages 10012--10022, 2021.

\bibitem[Lv et~al.(2024)Lv, Zhang, Qi, Li, and Huo]{lv2024lightweight1}
Cheng Lv, Enxu Zhang, Guowei Qi, Fei Li, and Jiaofei Huo.
\newblock A lightweight parallel attention residual network for tile defect recognition.
\newblock \emph{Scientific Reports}, 14\penalty0 (1):\penalty0 21872, 2024.

\bibitem[Mourya et~al.(2019)Mourya, Kant, Kumar, Gupta, and Gupta]{mourya2019all}
S Mourya, S Kant, P Kumar, A Gupta, and R Gupta.
\newblock All challenge dataset of isbi. 2019.
\newblock \emph{The Cancer Imaging Archive}, 2019.

\bibitem[Nam et~al.(2024)Nam, Syazwany, Kim, and Lee]{nam2024modality}
Ju-Hyeon Nam, Nur~Suriza Syazwany, Su~Jung Kim, and Sang-Chul Lee.
\newblock Modality-agnostic domain generalizable medical image segmentation by multi-frequency in multi-scale attention.
\newblock In \emph{Proceedings of the IEEE/CVF Conference on Computer Vision and Pattern Recognition}, pages 11480--11491, 2024.

\bibitem[Nickparvar(2021)]{nickparvar2021}
Msoud Nickparvar.
\newblock Brain tumor {MRI} dataset.
\newblock Data set, 2021.
\newblock Accessed on 3rd March.

\bibitem[Peng et~al.(2022)Peng, Wei, Deng, Wang, and Hu]{peng2022balanced}
Xiaokang Peng, Yake Wei, Andong Deng, Dong Wang, and Di Hu.
\newblock Balanced multimodal learning via on-the-fly gradient modulation.
\newblock In \emph{Proceedings of the IEEE/CVF conference on computer vision and pattern recognition}, pages 8238--8247, 2022.

\bibitem[Plissiti et~al.(2018)Plissiti, Dimitrakopoulos, Sfikas, Nikou, Krikoni, and Charchanti]{plissiti2018sipakmed}
Maria~E Plissiti, Panagiotis Dimitrakopoulos, Giorgos Sfikas, Christophoros Nikou, Orestis Krikoni, and Avraam Charchanti.
\newblock Sipakmed: A new dataset for feature and image based classification of normal and pathological cervical cells in pap smear images.
\newblock In \emph{2018 25th IEEE International Conference on Image Processing (ICIP)}, pages 3144--3148. IEEE, 2018.

\bibitem[Qin et~al.(2021)Qin, Zhang, Wu, and Li]{Qin_2021_ICCV}
Zequn Qin, Pengyi Zhang, Fei Wu, and Xi Li.
\newblock Fcanet: Frequency channel attention networks.
\newblock In \emph{Proceedings of the IEEE/CVF International Conference on Computer Vision (ICCV)}, pages 783--792, 2021.

\bibitem[Rahman and Marculescu(2023)]{rahman2023medical}
Md~Mostafijur Rahman and Radu Marculescu.
\newblock Medical image segmentation via cascaded attention decoding.
\newblock In \emph{Proceedings of the IEEE/CVF winter conference on applications of computer vision}, pages 6222--6231, 2023.

\bibitem[Rahman et~al.(2024)Rahman, Munir, and Marculescu]{rahman2024emcad}
Md~Mostafijur Rahman, Mustafa Munir, and Radu Marculescu.
\newblock Emcad: Efficient multi-scale convolutional attention decoding for medical image segmentation.
\newblock In \emph{Proceedings of the IEEE/CVF Conference on Computer Vision and Pattern Recognition}, pages 11769--11779, 2024.

\bibitem[Rahman et~al.(2020)Rahman, Khandakar, Kadir, Islam, Islam, Mazhar, Hamid, Islam, Kashem, Mahbub, et~al.]{rahman2020reliable}
Tawsifur Rahman, Amith Khandakar, Muhammad~Abdul Kadir, Khandaker~Rejaul Islam, Khandakar~F Islam, Rashid Mazhar, Tahir Hamid, Mohammad~Tariqul Islam, Saad Kashem, Zaid~Bin Mahbub, et~al.
\newblock Reliable tuberculosis detection using chest x-ray with deep learning, segmentation and visualization.
\newblock \emph{Ieee Access}, 8:\penalty0 191586--191601, 2020.

\bibitem[Ronneberger et~al.(2015)Ronneberger, Fischer, and Brox]{ronneberger2015u}
Olaf Ronneberger, Philipp Fischer, and Thomas Brox.
\newblock U-net: Convolutional networks for biomedical image segmentation.
\newblock In \emph{Medical image computing and computer-assisted intervention--MICCAI 2015: 18th international conference, Munich, Germany, October 5-9, 2015, proceedings, part III 18}, pages 234--241. Springer, 2015.

\bibitem[Sawyer-Lee et~al.(2016)Sawyer-Lee, Gimenez, Hoogi, and Rubin]{sawyer2016curated}
R Sawyer-Lee, F Gimenez, A Hoogi, and D Rubin.
\newblock Curated breast imaging subset of digital database for screening mammography (cbis-ddsm)[skup podataka].
\newblock \emph{The cancer imaging archive}, 2016.

\bibitem[Shen et~al.(2021)Shen, Cao, Chen, Zhang, Su, Wu, Huang, and Ji]{shen2021parallel}
Yunhang Shen, Liujuan Cao, Zhiwei Chen, Baochang Zhang, Chi Su, Yongjian Wu, Feiyue Huang, and Rongrong Ji.
\newblock Parallel detection-and-segmentation learning for weakly supervised instance segmentation.
\newblock In \emph{Proceedings of the IEEE/CVF International Conference on Computer Vision}, pages 8198--8208, 2021.

\bibitem[Steiner et~al.(2021)Steiner, Kolesnikov, Zhai, Wightman, Uszkoreit, and Beyer]{steiner2021train}
Andreas Steiner, Alexander Kolesnikov, Xiaohua Zhai, Ross Wightman, Jakob Uszkoreit, and Lucas Beyer.
\newblock How to train your vit? data, augmentation, and regularization in vision transformers.
\newblock \emph{arXiv preprint arXiv:2106.10270}, 2021.

\bibitem[Szegedy et~al.(2016)Szegedy, Vanhoucke, Ioffe, Shlens, and Wojna]{szegedy2016rethinking}
Christian Szegedy, Vincent Vanhoucke, Sergey Ioffe, Jon Shlens, and Zbigniew Wojna.
\newblock Rethinking the inception architecture for computer vision.
\newblock In \emph{Proceedings of the IEEE conference on computer vision and pattern recognition}, pages 2818--2826, 2016.

\bibitem[Tschandl et~al.(2018)Tschandl, Rosendahl, and Kittler]{tschandl2018ham10000}
Philipp Tschandl, Cliff Rosendahl, and Harald Kittler.
\newblock The ham10000 dataset, a large collection of multi-source dermatoscopic images of common pigmented skin lesions.
\newblock \emph{Scientific data}, 5\penalty0 (1):\penalty0 1--9, 2018.

\bibitem[Zheng et~al.(2023)Zheng, Liu, Qi, and Chen]{zheng2023potter}
Ce Zheng, Xianpeng Liu, Guo-Jun Qi, and Chen Chen.
\newblock Potter: Pooling attention transformer for efficient human mesh recovery.
\newblock In \emph{Proceedings of the IEEE/CVF Conference on Computer Vision and Pattern Recognition}, pages 1611--1620, 2023.

\end{thebibliography}

}

\end{document}